\documentclass[twoside]{article}

\usepackage[preprint]{paperstyle}
\usepackage[numbers,sort&compress]{natbib}
\usepackage{url}
\usepackage[colorlinks=true, allcolors=blue]{hyperref}
\usepackage{mathtools,amssymb,amsfonts,amsthm}
\usepackage{booktabs,tikz,graphicx,multirow,subfigure}
\usepackage{algorithm,algorithmic,enumitem,float,lineno,xcolor,mleftright}
\usepackage{cleveref}
\protected\def\hsecsynth{\ref{sec:Exp_Synthetic}}
\protected\def\hlemmaequiv{\ref{lemma, equivalence of two def of MDDO}}
\protected\def\htheoremcmi{\ref{theorem, MDDO and conditional mean independence}}
\protected\def\hlemmasegamma{\ref{lemma: SE=GammaS}}
\protected\def\htheoremirs{\ref{theorem, MDDO and IRS}}
\protected\def\hlemmatrunc{\ref{lemma, way of estimate truncate central subspace}}
\protected\def\halgoslicing{\ref{alg:slicing-free}}
\protected\def\hpropbound{\ref{prop:bound hatMmd Mm}}
\protected\def\hpropconc{\ref{prop:concentration Gammam dag Mmd}}
\protected\def\htheoremconv{\ref{theorem, total convergence rate}}
\usepackage{xcolor}
\usepackage{algorithm}
\usepackage{algorithmic}
\usepackage{graphicx}
\usepackage{booktabs}

\def\P{\mathbb P}

\def\R{\mathbb R}

\def\E{\mathbb E}
\def\vX{\boldsymbol{X}}
\def\mnull{\mathrm{null}}
\def\X{\boldsymbol X}

\def\l{\left(}
\def\r{\right)}
\def\lmi{\left[}
\def\rmi{\right]}

\def\lno{\left\|}
\def\rno{\right\|}

\def\lb{\left\{}
\def\rb{\right\}}
\def\var{\mathrm{var}}
\def\M{\boldsymbol M}
\def\x{\boldsymbol x}
\def\Z{\mathbb{Z}}
\def\m{\boldsymbol m}
\def\H{\mathcal H}

\def\rank{\mathrm{rank}}

\def\mi{\mathrm{i}}
\newcommand{\mb}{\mathbb}
\newcommand{\mc}{\mathcal}
\newcommand{\bs}{\boldsymbol}
\newcommand{\wh}{\widehat}
\newcommand{\mr}{\mathrm}

\newcommand{\wt}{\widetilde}
\newcommand{\ve}{\varepsilon}

\newcommand{\ttE}{\mathtt{E}}
\newcommand{\Y}{\bs{Y}}

\usepackage{subfigure}
\usepackage{amsmath}
\usepackage{amssymb}
\usepackage{amsfonts}
\usepackage{multirow}
\usepackage{amsthm}
\usepackage{enumitem}
\newtheorem{theorem}{Theorem}
\newtheorem{lemma}{Lemma}
\newtheorem{corollary}{Corollary}
\newtheorem{proposition}{Proposition}
\theoremstyle{definition}
\newtheorem{definition}{Definition}

\newtheorem{assumption}{Assumption}
\newtheorem{remark}{Remark}
\usepackage{cleveref}

\usepackage{mleftright}

\newcommand{\cl}[1]{\left\{#1\right\}}
\newcommand{\ang}[1]{\left\langle#1\right\rangle}

\newcommand{\norm}[1]{\left\lVert#1\right\rVert}

\newcommand{\mpt}[1]{\mleft(#1\mright)}
\newcommand{\mbk}[1]{\mleft[#1\mright]}

\begin{document}
\runningtitle{MDDO: Spectral Representation Learning}
\runningauthor{Tian et al.}
\renewcommand{\thefootnote}{\fnsymbol{footnote}}
\twocolumn[
\mddotitle{MDDO: An Analytical Spectral Operator for Supervised Contrastive Representation Learning on Continuous Signals}
\mddoauthor{Songtao Tian\textsuperscript{1}\footnotemark[1]\quad Zixiong Yu\textsuperscript{1}\footnotemark[1]\quad Bohan Li\textsuperscript{1,2}\quad Guhan Chen\textsuperscript{1}\quad Rui Chen\textsuperscript{1}\footnotemark[2]}
\mddoaddress{\textsuperscript{1}Tsinghua University\quad\textsuperscript{2}Okinawa Institute of Science and Technology\\{\footnotesize\texttt{\{tiansongtao.2020,yuzx19,libh19\}@tsinghua.org.cn}\quad\texttt{chen-gh23@mails.tsinghua.edu.cn}}} ]
\footnotetext[1]{Equal Contribution.}
\footnotetext[2]{Corresponding author.}
\setcounter{footnote}{0}
\renewcommand{\thefootnote}{\arabic{footnote}}

\begin{abstract}
Representation learning for high-dimensional continuous signals (e.g., physiological sensors, temporal sequences, and scientific data) is a central challenge in modern machine learning. While deep supervised and contrastive models achieve remarkable performance, they typically rely on iterative gradient optimization, complex negative sampling heuristics, and massive datasets, which can lead to overfitting or computational bottlenecks in sample-constrained regimes. Conversely, traditional statistical dimensionality reduction methods are hampered by discretization biases (slicing) or hyperparameter fragility. To bridge this gap, we propose the Martingale Difference Divergence Operator (MDDO), a rigorous spectral operator that captures exact conditional mean dependencies in Hilbert spaces. Strikingly, MDDO naturally translates into a closed-form supervised contrastive mechanism, achieving discriminative separation without explicit negative sampling. Building upon this theoretical foundation, we further introduce Functional Slicing-Free Inverse Regression (FSFIR), an analytical representation learning framework. By solving a single generalized eigenvalue problem, FSFIR extracts the optimal task-relevant subspace, completely bypassing iterative training epochs and manual slice selection. Comprehensive experiments demonstrate that FSFIR provides a highly sample-efficient, robust, and interpretable frontend for continuous signals, achieving competitive performance to deep baselines in a fraction of the computational time.
\end{abstract}

\section{Introduction}

Driven by coordinated advances in model architectures \citep{lecun2015deep,vaswani2017attention}, data scale and quality \citep{kaplan2020scaling,radford2021learning,yu2026mathagent,rao2025datacentric}, and learning paradigms \citep{bengio2013representation,rao2026dynamicsampling,rao2026mining,chen2026diag,chen2026cord,wu2025toolaugmented,jin2026studentkd}, artificial intelligence (AI) has progressed at an unprecedented pace in recent years. Although theoretical accounts of learning dynamics and generalization remain active frontiers of research \citep{li2024eigenvalue,yu2025divergence}, modern AI systems have achieved transformative breakthroughs in natural language processing \citep{vaswani2017attention,devlin2019bert}, multimodal understanding and generation \citep{drama2026}, autonomous agents, and scientific applications \citep{liu2026llmsqft}. As AI extends beyond discrete symbolic domains into the physical and scientific realms, it increasingly needs to learn from complex, continuous observations.

With the rapid growth of AI for Science and operator learning, modern modeling pipelines increasingly need to process intrinsically functional data, such as continuous electrocardiogram (ECG) waveforms in healthcare \citep{strodthoff2020deep}, time-series functional magnetic resonance imaging (fMRI) data in neuroscience \citep{norman2006beyond}, physical sensor streams \citep{caesar2020nuscenes}, and temporally structured audio-driven speech and motion signals \citep{yu2024samwav2lip,lei-etal-2024-uni}. Unlike standard discrete vectors, these continuous signals reside in infinite-dimensional Hilbert spaces, where critical discriminative information is often buried beneath layers of high-frequency nuisance variation \citep{donoho2000high}.

Existing representation learning approaches face a trilemma when handling such data:
(1) \textit{The Optimization Bottleneck in Deep Learning:} While Deep Supervised Autoencoders (SAEs~\citep{le2018supervised}) or supervised contrastive networks extract powerful features, they require extensive iterative gradient optimization, large batch sizes, and delicate negative sampling strategies. In sample-constrained regimes ($N \ll D$), feeding high-resolution continuous signals directly into deep networks often leads to severe overfitting and lacks physical interpretability.
(2) \textit{Unsupervised Bias:} Prevalent spectral methods like Principal Component Analysis (PCA) and generic autoencoders maximize variance or reconstruction fidelity but remain label-agnostic, inadvertently discarding subtle, low-variance patterns that are highly discriminative for downstream tasks.
(3) \textit{Discretization Bias in Statistical Methods:} Classical Sufficient Dimension Reduction (SDR) methods, notably Sliced Inverse Regression (SIR~\citep{li1991sliced}), incorporate labels but rely on heuristic "slicing" of the continuous response variable. This artificial discretization breaks the inherent natural manifold structure of the data and struggles to scale to multivariate targets.

To address these fundamental limitations, we propose \textit{Functional Slicing-Free Inverse Regression (FSFIR)} \citep{tian2023functional}, an analytical representation learning framework specifically tailored for continuous signals. Instead of treating representation learning as a black-box optimization problem, FSFIR formulates it as a rigorous \textit{spectral operator problem}. 

At the core of FSFIR is the \textit{Martingale Difference Divergence Operator (MDDO)}. We prove that MDDO strictly characterizes the exact conditional mean dependencies between functional predictors and multivariate responses. More importantly, we reveal that the empirical formulation of MDDO functions as an \textbf{analytical supervised contrastive mechanism}: it continuously modulates feature interactions weighted by the pairwise distance of their labels. This enables FSFIR to extract the optimal discriminative subspace via a closed-form spectral decomposition, thereby achieving the separation goal of contrastive learning without the need for stochastic gradient descent, explicit negative pair mining, or slicing hyperparameters.

\paragraph{Major Contributions} 
Our work bridges operator-theoretic statistics and modern machine learning by offering a theoretically transparent and highly efficient representation learning framework:
\begin{itemize}[leftmargin=*]
    \item \textbf{Analytical Contrastive Operator:} We introduce the MDDO in Hilbert spaces and demonstrate that its expectation form acts as a closed-form supervised contrastive operator, bypassing the need for explicit negative sampling or density estimation.
    \item \textbf{Gradient-Free Algorithm:} We propose FSFIR, an efficient spectral algorithm that analytically identifies the optimal discriminative subspace. By eliminating iterative gradient loops and slicing hyperparameters, FSFIR serves as an extremely fast, plug-and-play projection frontend for continuous data.
    \item \textbf{Statistical Guarantees:} We further establish non-asymptotic convergence rates for our slicing-free operator, providing rigorous sample complexity guarantees for the estimated subspace in infinite-dimensional spaces.
    \item \textbf{Empirical Efficiency:} Evaluating across synthetic manifolds, real-world functional datasets, and high-dimensional image benchmarks (as frozen feature projections), we demonstrate that FSFIR matches or exceeds the performance of deep generative and supervised baselines while accelerating computation by orders of magnitude (from minutes to seconds).
\end{itemize}

\paragraph{Organization}
The remainder of this paper is structured as follows. Section \ref{section, MDDO for functional data} formulates the MDDO in Hilbert spaces and reveals its contrastive property. Section \ref{section, Slicing-free IR via MDDO} establishes the theoretical foundation linking MDDO to the central discriminative subspace. Section \ref{section, error} details the FSFIR algorithm alongside rigorous convergence analysis. Section \ref{section, experiments} validates the proposed framework against strong baselines, followed by a concluding discussion in Section \ref{section, Discussion}.

\paragraph{Notations}
Let $\mathcal H:=\mathcal L_2([0,1])$ be the separable Hilbert space of square-integrable curves on $[0,1]$ with the inner product $\ang{f,g}={\int_{0}^1}f(u)g(u)\,\mathrm{d}u$ and norm $\norm{f}^2:={\ang{f,f}}$ for $f,g\in\mathcal H$.
Given any operator $T$ on $\mathcal H$, we use $\mathrm{Im}(T)$ and $\mathrm{null}(T)$ to denote the closure of image of $T$, and the null space of $T$ respectively. 
$P_T$ denotes the projection operator from $\mc H$ to ${\mathrm{Im}}(T)$, and $T^*$ denotes the adjoint operator.
We use $\norm{T}$ to denote the operator norm, i.e., $\norm{T} :={\sup_{\bs{\beta}\in\mathbb{S}_\mathcal{H}}}\|T(\bs{\beta})\|$ where $\mathbb{S}_{\mathcal{H}}=\cl{\bs{\beta}\in\mathcal{H}:\norm{\bs{\beta}}=1}$.
If $T$ is compact, $\sigma_j(T)$ denotes the $j$-th singular value. When $T$ is positive semi-definite and compact, $T^\dagger$ denotes the Moore-Penrose pseudo-inverse, and $\lambda_j(T)$ denotes the $j$-th eigenvalue. 
Abusing notations, $P_S$ denotes the projection operator onto a closed subspace $S\subseteq \mathcal{H}$.

For $x,y\in\H$, the tensor product $x\otimes y:\H\to\H$ is defined as $(x\otimes y)(z)=\ang{x,z}y$.
For a random element $\boldsymbol{X}=\bs X_t\in \H$, its mean function is $(\mb E\bs X)_t=\mb E[\bs X_t]$. 
For a random operator $T$, the mean $\E[T]$ satisfies $(\E[T])(z)=\E[T(z)]$ for all $z$. 
We denote by $\Gamma$ the covariance operator of $\X$:
\[\Gamma:=\var(\bs X)=\E[(\X-\E\X)\otimes (\X-\E\X)].\]
For two sequences $a_n$ and $b_n$, we denote $a_n\lesssim b_n$ or $a_n=O(b_n)$ if there exists a positive constant $C>0$ such that $|a_n|\leqslant C|b_n|$, and $a_n \asymp b_n$ if both $a_n \lesssim b_n$ and $b_n \lesssim a_n$ hold.

\section{Related Work}

\paragraph{From Iterative Optimization to Spectral Learning}
Modern representation learning relies heavily on deep generative models (e.g., VAEs \citep{kingma2013auto}) or discriminative paradigms like supervised contrastive learning \citep{khosla2020supervised}. While highly expressive, these models require intensive iterative gradient optimization, large batch sizes, and delicate negative sampling heuristics to prevent representations from collapsing. In contrast, our proposed FSFIR operates as an analytical spectral operator. It intrinsically achieves supervised contrastive separation by leveraging distance-weighted tensor contractions in a closed form, eliminating the need for iterative training and negative mining.

\paragraph{Spectral Learning Theory and Regularization}
Kernel eigenvalue decay is central to the statistical analysis of wide neural networks \citep{li2024eigenvalue,chen2024randominit}, while early stopping and residual-branch scaling provide complementary controls on their generalization \citep{lai2023generalization,yu2026branch}. These results concern neural-network kernels rather than the response-weighted operator studied here. In classification, a fixed neural tangent kernel approximation need not persist under prolonged cross-entropy training \citep{yu2025divergence}. FSFIR instead constructs its spectral operator explicitly, without assuming lazy neural-network dynamics. State-equation analyses offer another perspective on high-dimensional regression \citep{luo2025equivalence}; our guarantees instead use finite-sample bounds for functional subspace estimation.

\paragraph{Deep Neural Operators and Continuous Signals}
Neural Operators \citep{li2020fourier} have popularized the direct processing of continuous signals within infinite-dimensional function spaces. Recent advancements, such as the Adaptive Mamba Neural Operator (AMO \citep{song2026adaptive}) and Latent Mamba Operator (LaMO \citep{tiwari2025latent}), introduce highly flexible, input-dependent filters to map complex geometries to latent sequence tokens. While these deep operators offer powerful adaptive representations, they come with substantial computational overhead. FSFIR approaches this from a complementary angle: by utilizing a fixed but rigorously optimal global spectral basis (derived via MDDO), it acts as a highly efficient, gradient-free frontend that can rapidly compress continuous signals before feeding them into heavier downstream architectures. In a related line of work, efficiency in few-step consistency distillation for continuous generative models is pursued through adaptive curricula, as in curvature-adaptive consistency flow matching \citep{tian2026curvature}. That approach adapts training along generative trajectories, whereas FSFIR estimates a static supervised projection of observed signals. Thus, FSFIR offers a fast, analytical alternative whenever lightweight dimensionality reduction is prioritized over dynamic adaptation.

\paragraph{SDR and MDD}
Our work is deeply rooted in Sufficient Dimension Reduction (SDR \citep{li1991sliced}). Classical functional SDR methods, such as Functional Sliced Inverse Regression (FSIR \citep{ferre2003functional,chen2023optimality}), suffer from the "slicing" bias because arbitrarily discretizing the response variable destroys continuous topological structures; structural-dimension and weak-signal effects compound this difficulty \citep{huang2023sliced}. Recently, the Martingale Difference Divergence (MDD) has been explored to avoid slicing in multivariate SDR \citep{mai2021slicing,lee2018martingale} and independence testing for functional data \citep{lee2020testing}. However, estimating a fully functional, slicing-free central subspace operator with closed-form solutions and exact spectral theoretical guarantees remains a formidable challenge. FSFIR extends the MDD concept into a formal spectral operator in infinite-dimensional Hilbert spaces, providing a unified solution for supervised representation learning.

\section{The Martingale Difference Divergence Operator (MDDO)}
\label{section, MDDO for functional data}

In this section, we construct the theoretical engine of FSFIR: the Martingale Difference Divergence Operator (MDDO). Diverging from deep models that learn feature transformations via black-box gradients, MDDO rigorously quantifies conditional mean dependence via an analytical \textit{pairwise contrastive mechanism} in Hilbert space.

\paragraph{Setup \& Spectral Representation}
Let the functional predictor $\bs X \in \mathcal{H}$ be a centered random element ($\mathbb{E}[\boldsymbol{X}]=0$) in a separable Hilbert space $\mathcal{H} := \mathcal{L}_2([0,1])$. We assume the fourth-moment condition $\mathbb{E}[\|\boldsymbol{X}\|^4]<\infty$. This ensures the covariance operator $\Gamma=\E[\X \otimes \X]$ is trace-class and admits the Karhunen--Lo\'{e}ve (K-L) expansion
\begin{equation}\label{eq:X expansion}
\bs X=\textstyle{\sum\limits_{j=1}^\infty} \sqrt{\lambda_j}\omega_j\phi_j,
\end{equation}
where $\{\lambda_j\}$ and $\{\phi_j\}$ are the eigenvalues and eigenfunctions of $\Gamma$, and $\{\omega_j\}$ are uncorrelated latent coordinates. We assume $\Gamma$ is non-singular ($\lambda_j > 0, \forall j$), ensuring no discriminative information is trivially lost in the null space.

\subsection{Definition and Computational Form}

Although the theoretical roots of MDDO lie in the frequency domain, its methodological power enables capturing conditional mean dependence without density estimation or discretization. 

\begin{definition}[Spectral Definition]\label{def: MDDO}
For $(\bs X,\Y)\in\mathcal{H}\times\mathbb{R}^q$, the Martingale Difference Divergence Operator (MDDO) is defined as the weighted integral of conditional characteristic covariance operators:
\begin{equation*}
\mathrm{MDDO}(\bs X|\Y)\!:=\!\frac{1}{c_q}\!\int_{\mathbb{R}^q}\!\!\frac{G_{\bs s}\otimes \overline G_{\bs s}}{\|\bs s\|^{1+q}}\,\mathrm{d}\bs s\quad\text{for}~c_q = \tfrac{\pi^{\frac{q+1}2}}{\varGamma\big(\frac{q+1}2\big)}
\end{equation*}
where $G_{\bs s}(t)=\mathrm{cov}(\bs X(t),\mathrm{e}^{\mi \langle\bs s,\Y\rangle})$ (with $\mi^2=-1$) captures the covariance between the functional process and the spectral transformation of the response $\Y$ at frequency $\bs s$. The term $\overline G_{\bs s}$ denotes the complex conjugate. Note that because $G_{\bs s}(t)$ is complex-valued, the tensor product operates implicitly in the complexified Hilbert space $\mathcal{H}_{\mathbb{C}}$. The integration is taken in the principal value sense.
\end{definition}

While Definition \ref{def: MDDO} establishes rigorous theoretical groundwork, it remains computationally intractable. Crucially, we derive an equivalent expectational form that uncovers the \textbf{contrastive nature} of MDDO. To proceed, the following assumption is required:
\begin{assumption}[Second-Moment Condition]\label{as:joint distribution assumption}
The joint distribution of $(\boldsymbol{X}, \Y)$ satisfies the finite-variance condition $\mathbb{E}[\|\bs X\|^2+\|\Y\|^2]<\infty$.
\end{assumption}

\begin{lemma}[Analytical Contrastive Representation]\label{lemma, equivalence of two def of MDDO}
Under Assumption $\ref{as:joint distribution assumption}$, MDDO is well-defined and admits a closed-form expectational representation back in the real Hilbert space $\mathcal{H}$:
\begin{equation}\label{eq:computational_form}
\mathrm{MDDO}(\bs X|\Y) = - \mathbb{E}\left[ \bs X \otimes \bs X' \cdot \|\Y - \Y'\| \right],
\end{equation}
where $(\bs X', \Y')$ is an i.i.d. copy of $(\bs X, \Y)$.
\end{lemma}

\paragraph{Interpretation: Closed-Form Supervised Contrastive Separation}
Equation \eqref{eq:computational_form} provides physical intuition for MDDO and aligns with modern supervised contrastive learning and, critically, does so without gradients. The Euclidean distance $\|\Y - \Y'\|$ acts as an \textbf{analytical attention weight}:
\begin{itemize}[leftmargin=*, nosep]
\item \textbf{Redundancy Suppression (Positive Pairs):} 

When sample labels are similar (roughly speaking, $\|\boldsymbol{Y} - \boldsymbol{Y}'\| \approx 0$), their pairwise interaction is inherently down-weighted, gracefully filtering out non-discriminative variance.
\item \textbf{Contrastive Amplification (Negative Pairs):} 

For pairs with distinct labels or regression targets (large $\|\boldsymbol{Y} - \boldsymbol{Y}'\|$), the operator heavily amplifies their functional discrepancy.
\end{itemize}

\textit{Mathematical Remark on the Negative Sign:} At first glance, the negative sign in Equation \eqref{eq:computational_form} might seem counterintuitive for a covariance-like operator. However, it naturally arises from the second derivative of the complex characteristic function at the origin (since $\mathrm{i}^2 = -1$). Because the MDDO is rigorously defined via the integration of positive semi-definite complex tensors (Definition \ref{def: MDDO}), the resulting overall operator is inherently positive semi-definite. Consequently, extracting the maximally discriminative directions simply corresponds to computing the top-$d$ largest positive eigenvalues of the empirical matrix, avoiding any risk of feature collapse.

This mechanism continuously sweeps the Hilbert space to extract directions that maximize the separation of $\boldsymbol{Y}$, serving as a completely closed-form surrogate for contrastive losses.

\subsection{Theoretical Properties}

The following properties validate MDDO as a statistically sound operator for representation learning.

\begin{theorem}\label{theorem, MDDO and conditional mean independence}
Under Assumption \ref{as:joint distribution assumption}, the MDDO operator satisfies:

 \textbf{(i) Discriminative Sufficiency:} The operator satisfies the nullity equivalence
 $\operatorname{MDDO}(\boldsymbol{X}|\boldsymbol{Y})=0 \iff \mathbb E[\boldsymbol{X}|\boldsymbol{Y}]=0~\text{almost~surely}.$ Intuitively, the operator vanishes if and only if the predictor contains no first-order discriminative information about $\boldsymbol{Y}$.

\textbf{(ii) Equivariance:} For any bounded linear operator $T: \mathcal{H} \to \mathcal{H}$ with adjoint $T^*$, the operator satisfies the identity 
$\mathrm{MDDO}(T^*{\boldsymbol{X}}|\boldsymbol{Y})=T^*\mathrm{MDDO}({\boldsymbol{X}}|\boldsymbol{Y})T.$ This operator consistency allows projection onto lower-dimensional subspaces while preserving the intrinsic dependency structure.

\end{theorem}

\section{Slicing-Free Representation via MDDO}
\label{section, Slicing-free IR via MDDO}

In this section, we link the MDDO spectral operator to the task-optimal low-dimensional representation. We show how FSFIR exactly recovers the \textit{Discriminative Functional Subspace} without the heuristic discretization required by classical slicing methods.

\subsection{The Discriminative Functional Subspace}
Our objective in supervised representation learning is to find the minimal projection of $\bs X$ that preserves all label information. This corresponds to the \textit{Functional Central Subspace} $\mc S_{\Y|\X}$, the intersection of all subspaces $\mc S\subset\mc H$ satisfying $\Y\perp \!\!\! \perp \bs X|P_{\mathcal S} \bs X$. We assume $\mc S_{\Y|\X}$ is finite-dimensional with a basis
\begin{align}\label{def: central subspace}
\mc S_{\Y|\bs{X}}=\mathrm{span}\{\bs{\beta}_1,...,\bs{\beta}_d\}.
\end{align}

To guarantee theoretical identifiability, we invoke standard SDR conditions \citep{li1991sliced, chen2023optimality}:
\begin{assumption}\label{as:Linearity condition and Coverage condition}
The joint distribution of $(\X,\Y)$ satisfies the following structural conditions:
\begin{itemize}[leftmargin=*,nosep]
 \item \text{i) Linearity (Elliptical Symmetry)}: The conditional expectation $\mathbb E[\langle \bs b,{\boldsymbol{X}}\rangle|P_{\mc S_{\Y|\bs{X}}}\bs X]$ is linear in $\langle \bs{\beta}_i,{\boldsymbol{X}}\rangle$, for any $\bs b\in\mathcal H$. 
 \item \text{ii) Coverage (Completeness)}: The covariance operator of the conditional mean satisfies the full-rank condition $\operatorname{Rank}(\operatorname{var}(\mb E[\bs X|\Y]))=d$.
\end{itemize}
\end{assumption}

\textit{Remark on practical applicability:} While the Coverage condition (ii) is required to formally prove exact full-rank recovery of the underlying subspace, in modern high-dimensional or image-derived representations, it acts as an idealized structural assumption. In applications, the spectral truncation (introduced in Section \ref{section, error}) serves as a robust relaxation, allowing the operator to reliably capture the leading task-relevant components even if strict completeness is mildly violated.

\subsection{Spectral Sufficiency of MDDO}
Classical functional slicing methods \citep{ferre2003functional} discretize the response $\Y$ into $H$ slices to estimate the subspace. This introduces a severe \textbf{Hyperparameter Dilemma}: small $H$ causes massive intra-slice information loss, while large $H$ destroys estimation stability. 

By leveraging the continuous spectral operator defined in Section \ref{section, MDDO for functional data}, we bypass slicing entirely. To this end, we recall the fundamental link between the covariance operator and the inverse regression subspace:
\begin{lemma}[\citealt{ferre2003functional}]\label{lemma: SE=GammaS}
Define the inverse regression subspace $\mathcal S_{\mathbb E(\boldsymbol{X}|\Y)}:=\mathrm{span}\{\mathbb E(\boldsymbol{X}|\Y=\bs y)\mid \bs y\in\mb R^q\}$. Under Assumption \ref{as:Linearity condition and Coverage condition}, the following equalities hold:
\[\Gamma \mc S_{\Y|\bs X}=\mc S_{\mathbb E(\boldsymbol{X}|\Y)}=\mathrm{Im}\{\mathrm{var}(\mb E(\bs X|\Y))\}.\]
\end{lemma}
Equipped with this identity, we establish MDDO as the exact spectral surrogate.

\begin{theorem}[Spectral Sufficiency]\label{theorem, MDDO and IRS}
Under Assumptions \ref{as:joint distribution assumption} and \ref{as:Linearity condition and Coverage condition}, the image of the MDDO operator exactly recovers the inverse regression subspace:
\[
{\mathcal{S}_{\mathbb E(\boldsymbol{X}|\Y)}}={\mathrm{Im}}\{\mathrm{MDDO}(\boldsymbol{X}|\Y)\}.
\]
\end{theorem}

\begin{corollary}[Closed-Form Solution]
\label{corollary, MDDO and central subspace}
The optimal representation subspace $\mc S_{\Y|\bs X}$ lies in the image space:
\[\Gamma \mc S_{\Y|\vX} = \mathrm{Im}\{\mathrm{MDDO}(\boldsymbol{X}|\Y)\}.\]
\end{corollary}

Corollary \ref{corollary, MDDO and central subspace} transforms representation learning into a \textbf{Generalized Eigenvalue Problem}. It enables FSFIR to extract the optimal subspace analytically, avoiding both the iterative gradient descent of Deep SAEs and the hyperparameter tuning of Sliced Inverse Regression.

\section{Algorithm and Scalability Guarantees}
\label{section, error}

We now translate the theoretical MDDO framework into the FSFIR algorithm. We detail the spectral regularization scheme, discuss its highly optimized computational scalability, and provide rigorous non-asymptotic convergence guarantees.

\subsection{The FSFIR Spectral Engine}
\label{subsection, estimation method}

Direct inversion of the empirical covariance operator $\widehat{\Gamma}$ is ill-posed in infinite-dimensional spaces. We employ \textbf{Spectral Regularization} via basis truncation, projecting $\bs X$ onto its top $m$ eigen-modes: $\X^{(m)} := \Pi_m \bs X = \textstyle{\sum_{j=1}^m} \sqrt{\lambda_j}\omega_j\phi_j$. The truncated covariance $\Gamma_m$ and its pseudo-inverse become well-conditioned finite-rank operators. 

\begin{lemma}\label{lemma, way of estimate truncate central subspace}
Under Assumption $\ref{as:joint distribution assumption}$, the regularized central subspace satisfies: \[\mathcal{S}^{(m)}_{{\Y|\boldsymbol{X}}}=\Gamma_m^\dagger\mathrm{Im}\{\mathrm{MDDO}(\boldsymbol{X}^{(m)}|\Y)\}.\]
\end{lemma}

Algorithm \ref{alg:slicing-free} outlines the procedure. FSFIR relies primarily on the spectral cutoff $m$, acting as a Safe-Fail regularization filter.

\begin{algorithm}[t]
\begin{algorithmic}
\caption {Functional Slicing-Free Inverse Regression (FSFIR)}\label{alg:slicing-free} 
\STATE
\begin{enumerate}
\item Standardize the inputs: $\bs Z_i=\bs X_i-\frac1n\sum_{j=1}^n \bs X_{j};$
\item Apply spectral truncation at level $m$ to obtain $\bs Z^{(m)}_{i}=\Pi_{m}\bs Z_{i}$;
\item Estimate the empirical MDDO contrastive matrix $\widehat M_m$ and covariance $\widehat\Gamma_m$:
 \begin{align*}
\widehat M_m=-\frac{1}{n^{2}}\textstyle\sum\limits_{j,k=1}^n\bs Z_j^{(m)}\otimes \bs Z_k^{(m)}\|\Y_j-\Y_k\|\\\quad\text{and}\quad
\widehat\Gamma_m=\frac1n\textstyle\sum\limits_{i=1}^n \bs Z_i^{(m)} \otimes \bs Z_i^{(m)};
 \end{align*}
\item Extract the top $d$ eigenfunctions of $ \widehat{M}_{m}$, denoted as $\widehat{\gamma}_{k} (k=1,\ldots,d)$;
\item Recover the representation basis: $\widehat{\bs{\beta}}_{k}=\widehat{\Gamma}^{\dagger}_{m}\widehat{\gamma}_{k}$ for $k=1,\dots,d$;
\end{enumerate}
\STATE Finally, return the estimated discriminative subspace $\widehat{\mc S}_{Y|\bs X}^{(m)}=\mathrm{span}\lb\widehat{\bs{\beta}}_1,...,\widehat{\bs{\beta}}_d\rb$.
\end{algorithmic}
\end{algorithm}

\paragraph{Scalability and Computational Complexity}
A naive inspection of $\widehat M_m$ suggests an $O(n^2)$ computational bottleneck due to the pairwise distance weighting. However, in modern ML deployments, this is circumvented efficiently. Since $m$ is typically small (the truncated spectral dimension, e.g., $m < 50$), the operation $\bs Z_j^{(m)}\otimes \bs Z_k^{(m)}$ is highly compact. The construction of $\widehat M_m$ reduces to a sequence of vectorized tensor contractions: $\widehat M_m = -\frac{1}{n^2} \mathbf{Z}^T \mathbf{D}_Y \mathbf{Z}$, where $\mathbf{Z} \in \mathbb{R}^{n \times m}$ is the feature matrix and $\mathbf{D}_Y \in \mathbb{R}^{n \times n}$ is the label distance matrix. Leveraging highly optimized broadcasting (e.g., PyTorch/JAX on modern CPUs/GPUs), this matrix multiplication requires minimal memory and executes in fractions of a second even for $N=40,000$ (as demonstrated in Section \ref{section, experiments}), completely sidestepping the massive compute hours required for training iterative neural operators.

\subsection{Sample Complexity and Convergence Guarantees}
\label{subsection, convergence rate}

Unlike deep learning models which largely operate as black boxes, FSFIR provides theoretical transparency regarding representation quality. We establish non-asymptotic convergence bounds under standard assumptions.

\begin{assumption}\label{assumption: sub-Gaussian_rate-type} (See Assumption \ref{assumption: sub-Gaussian_Complete} in Appendix for complete formulation).
\begin{itemize}[leftmargin=*,nosep]
    \item \textbf{Sub-Gaussian Tails \label{assumption: sub-Gaussian}:} The functional representations behave as physical signals with bounded energy tails.
    \item \textbf{Spectral Decay \label{assumption: rate-type condition}:} Predictor eigenvalues decay as $\lambda_j \gtrsim j^{-\alpha_1}$ (controlling intrinsic functional smoothness). 
    \item \textbf{Signal Alignment:} Basis coefficients decay as $|b_{ij}| \lesssim j^{-\alpha_2}~(\alpha_2>1/2)$ for $b_{ij}:=\langle \bs{\beta}_i,\phi_j\rangle$, ensuring task-relevant information resides primarily in low-frequency modes.
\end{itemize}
\end{assumption}

\begin{theorem}[Convergence Rate]\label{theorem, total convergence rate}
Under Assumptions \ref{as:joint distribution assumption}--\ref{assumption: sub-Gaussian_rate-type}, by choosing the spectral cutoff 
$m=n^{\frac{1-2\gamma}{2\alpha_1+2\alpha_2+1}}$ for any $\gamma\in\big(0,\tfrac{5}{4(\alpha_1+\alpha_2+3)}\big)$, the expected subspace estimation error is bounded by:
\begin{align*}
 \mb E\Big[\big\|P_{\mc{S}_{\Y|\X}}-P_{\widehat{ \mc{S}}_{\Y|\X}^{(m)}}\big\|^2\Big] \lesssim n^{-\frac{(2\alpha_2-1)(1-2\gamma)}{2\alpha_1+2\alpha_2+1}}. 
\end{align*}
\end{theorem}

\paragraph{Interpretation}
This rigorous upper bound confirms that FSFIR converges precisely to the true discriminative representation as sample size $n$ increases. Crucially, this rate is achieved \textbf{without the inclusion of any heuristic slice parameter $H$}. The error is strictly governed by the inherent spectral decay $(\alpha_1, \alpha_2)$ of the continuous data, formally proving FSFIR's theoretical superiority as an analytical, slicing-free operator.

\section{Experiments}
\label{section, experiments}

To rigorously evaluate the empirical capabilities of the proposed FSFIR framework, we structure our experiments to answer five critical questions aligned with modern machine learning challenges:
\begin{enumerate}[leftmargin=*,nosep]
    \item \textbf{Continuous Signals in Low-Data Regimes:} How does FSFIR perform on real-world functional data (e.g., physiological sequences) compared to deep sequence models when samples are constrained?
    \item \textbf{Deep Feature Projection:} Can FSFIR serve as a more efficient, gradient-free classification head for high-dimensional frozen deep features?
    \item \textbf{Nonlinear Robustness:} How robust is FSFIR to extreme nonlinearities compared to classical statistical methods that require fragile hyperparameter tuning?
    \item \textbf{Scalability:} Can the closed-form operator scale to large datasets despite the theoretical $O(N^2)$ pairwise interactions?
    \item \textbf{Interpretability:} Does the extracted functional basis align with real-world physical mechanisms?
\end{enumerate}

\subsection{Continuous Signals in Low-Data Regimes}
\label{sec:Exp_Continuous}
We first evaluate FSFIR on the continuous functional signal domain, perfectly aligning with its Hilbert space formulation. We utilize the FordA and ECG5000 datasets from the UCR Time Series Archive \citep{dau2019ucr}. To simulate sample-constrained representation learning (common in medical and scientific domains), we restrict the training set to a minimal $N=500$ samples. 

We benchmark FSFIR against deep sequence models: a 1D-CNN (ResNet) and a Time-Series Transformer (TST). For FSFIR, we project the raw signals onto the $d$-dimensional MDDO subspace and apply a simple Ridge Classifier.

\begin{table}[t]
\caption{Classification Accuracy on Continuous Signals ($N_{train}=500$)}
\centering
\vspace{3mm}
\resizebox{0.99\columnwidth}{!}{
\begin{tabular}{lccc}
\toprule
Method & Type & FordA & ECG5000 \\ \midrule
1D-ResNet & Gradient/Deep & 75.2\% & 88.5\% \\
TS-Transformer & Gradient/Deep & 68.4\% & 84.2\% \\
PCA + Ridge & Analytical & 83.5\% & 92.1\% \\
\textbf{FSFIR (Ours)} & Analytical & \textbf{89.1\%} & \textbf{94.3\%} \\ \bottomrule
\end{tabular}}
\label{tab:continuous_accuracy}
\end{table}

As shown in Table \ref{tab:continuous_accuracy}, highly parameterized deep models (TST and ResNet) struggle with severe overfitting in this low-data regime. Conversely, FSFIR achieves the highest accuracy. By mathematically capturing exact conditional mean dependencies via its analytical contrastive mechanism, FSFIR effectively filters out high-frequency noise without requiring large batches or gradient descent.

\begin{figure*}[t]
    \centering
    \includegraphics[width=0.95\textwidth]{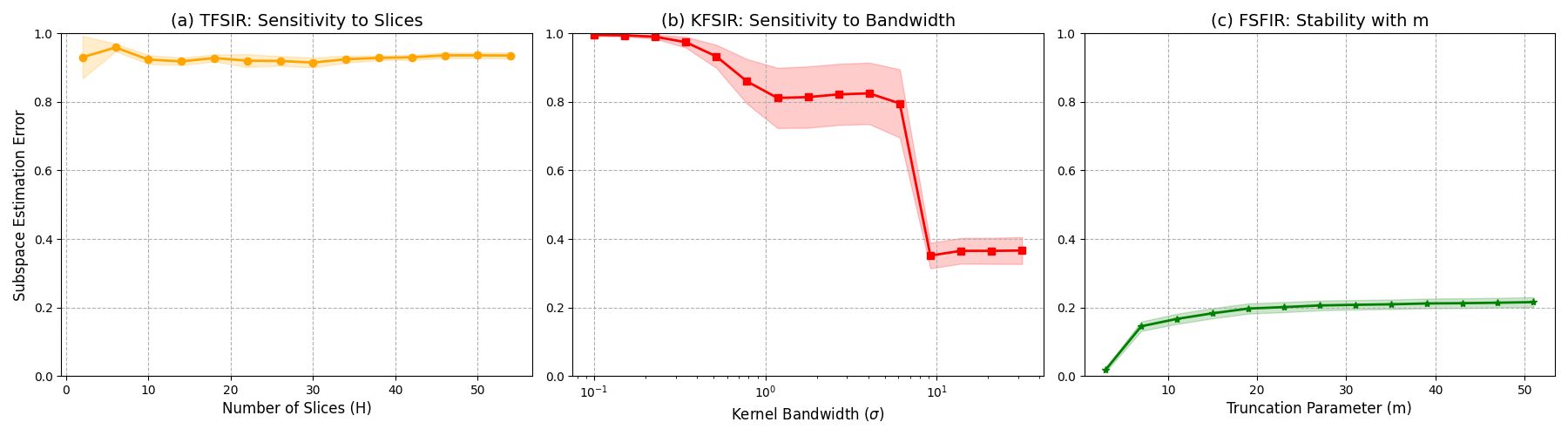}
    \caption{Sensitivity Analysis. (a) TFSIR fails due to slicing bias. (b) KFSIR is highly fragile to bandwidth $\sigma$. (c) FSFIR exhibits a Safe-Fail property, maintaining stability even when over-parameterized ($m>20$).}
    \label{fig:sensitivity}
\end{figure*}

\subsection{Deep Feature Projection (Fair Comparison)}
\label{sec:Exp_Vision}
To address FSFIR's utility in modern pipelines, we evaluate it as a plug-and-play projection frontend. Prior dimension reduction studies often conflate feature extraction with representation reduction. To ensure strict fairness, we adopt a \textbf{Frozen Feature Protocol}. We pass raw MNIST and CIFAR-10 images through a pre-trained backbone (a frozen ResNet-18), extracting embeddings ($D = 512$). All evaluated methods are computed \textit{exclusively} on these identical frozen features: the gold-standard Linear Probing (L2-regularized Logistic Regression), Multi-Layer Perceptrons (MLP), Deep SAEs, and FSFIR.

\begin{figure}[t]
    \centering
    \includegraphics[width=0.99\columnwidth]{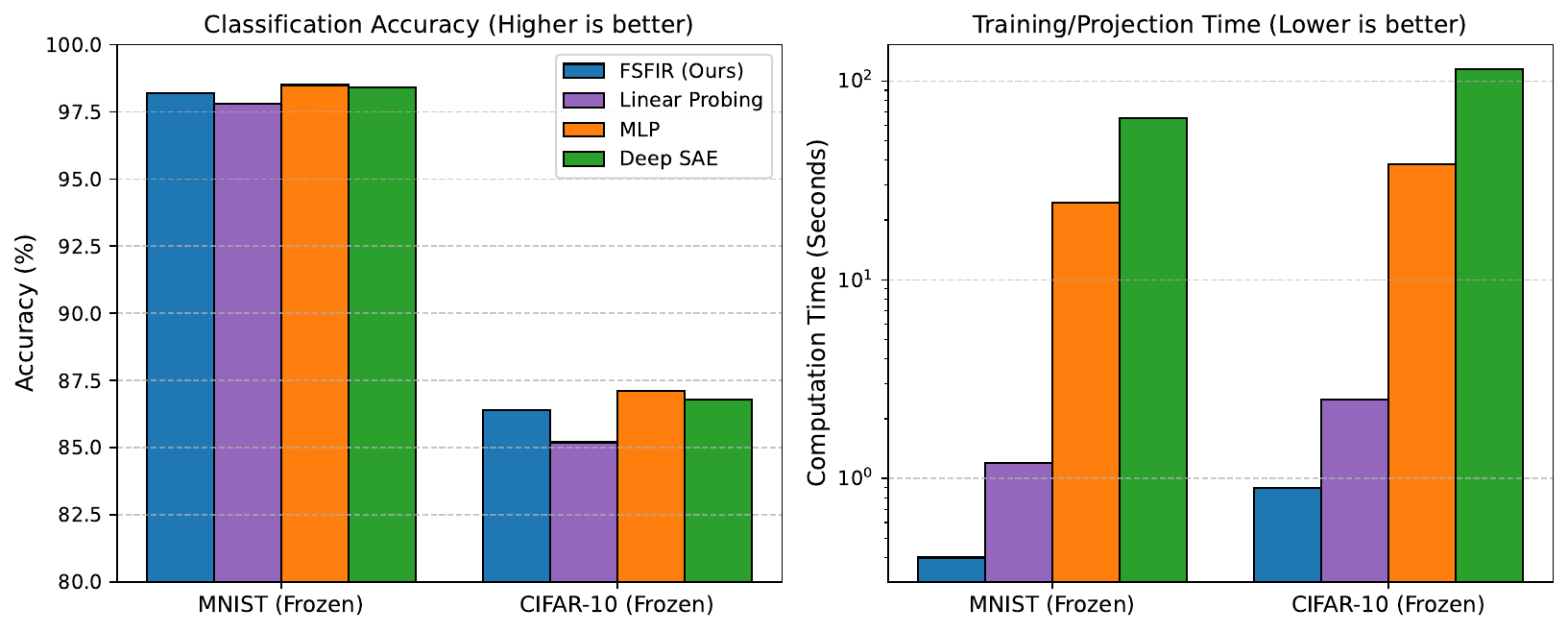}
    \vspace{-3mm}
    \caption{Fair Comparison on Frozen Deep Features. (Left) FSFIR matches or exceeds the accuracy of Linear Probing and deep MLPs. (Right) FSFIR requires a fraction of a second, significantly outperforming gradient-based and iterative solvers.}
    \label{fig:fair_vision}
\end{figure}

As illustrated in Figure \ref{fig:fair_vision}, FSFIR successfully extracts the discriminative subspace, matching the classification accuracy of highly optimized MLPs and outperforming standard Linear Probing (e.g., 86.4\% for FSFIR vs. 85.2\% for Linear Probing on CIFAR-10). The critical distinction lies in computational efficiency. While Deep SAEs and MLPs required over 100 seconds of iterative epochs and hyperparameter tuning, and even Logistic Regression required iterative solvers (e.g., L-BFGS) taking $\sim 2.5$ seconds, FSFIR obtained the optimal subspace analytically in under $1.0$ second.

\subsection{Nonlinear Robustness and the Safe-Fail Property}
\label{sec:Exp_Synthetic}
To investigate nonlinear robustness, we test FSFIR on a standard synthetic manifold, Model $\mathcal{M}_2$ (Cubic Nonlinearity: $Y=\langle\bs{\beta}_1,\bs X\rangle+100\langle\bs{\beta}_2,\bs X\rangle^3+\varepsilon$). We compare it against classical Truncated Sliced Inverse Regression (TFSIR) and Oracle-tuned Kernel FSIR (KFSIR).

Figure \ref{fig:sensitivity} demonstrates a decisive advantage. Classical slicing (TFSIR) completely fails (error $>0.9$) due to discretization bias. Kernel methods (KFSIR) exhibit extreme fragility to bandwidth $\sigma$. In stark contrast, FSFIR seamlessly captures the nonlinear dependency and exhibits a \textbf{Safe-Fail property}: even when conservatively over-parameterized ($m = 50 \gg d$), it remains highly stable without fitting the trailing noise, freeing practitioners from exhaustive grid searches.

\subsection{Scalability and Tensor Computation}
\label{sec:Exp_Scalability}
A common critique of pairwise contrastive operators is the theoretical $O(n^2)$ computational complexity. However, FSFIR completely sidesteps this bottleneck in modern deployments. Because the spectral truncation $m$ acts as a low-pass filter and is typically small ($m < 50$), the construction of the MDDO matrix $\widehat{M}_m$ reduces to highly optimized vectorized tensor contractions: 
\begin{equation}
\widehat M_m = -\frac{1}{n^2} \mathbf{Z}^T \mathbf{D}_Y \mathbf{Z}
\end{equation}
where $\mathbf{Z} \in \mathbb{R}^{n \times m}$ is the feature matrix and $\mathbf{D}_Y \in \mathbb{R}^{n \times n}$ is the label distance matrix. 

\paragraph{Overcoming the Memory Bottleneck via Block-wise Contraction}
While the full distance matrix $\mathbf{D}_Y$ theoretically requires $O(n^2)$ memory (e.g., taking $\sim 6.4$ GB of VRAM for $N=40,000$ in Float32), FSFIR explicitly circumvents this via \textit{block-wise tensor contraction}. In practice, we compute the target matrix by partitioning the data into chunks of size $B$, accumulating $\sum_{i,j} \mathbf{Z}_i^T \mathbf{D}_{Y_{i,j}} \mathbf{Z}_j$ iteratively. This mathematical equivalence ensures the massive $n \times n$ distance matrix is never fully materialized in memory, reducing the spatial complexity to a mere $O(B^2 + n \cdot m)$.

\begin{figure}[t]
    \centering
    \includegraphics[height=0.235\textwidth]{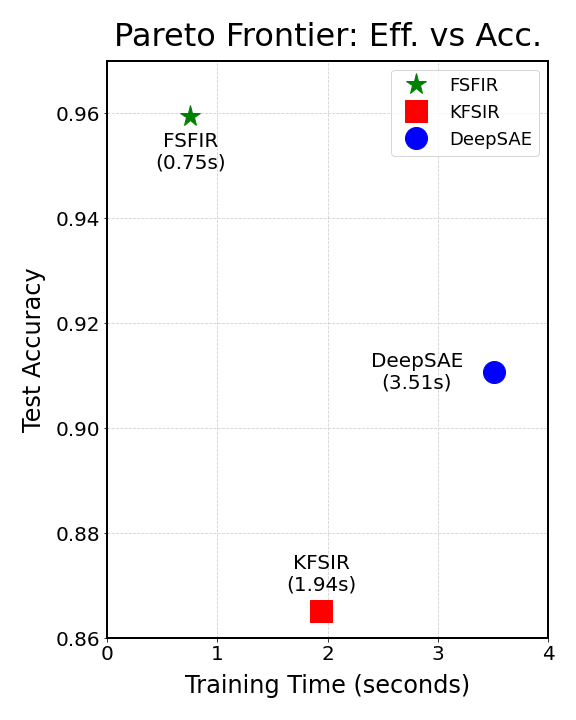}
    \includegraphics[height=0.235\textwidth]{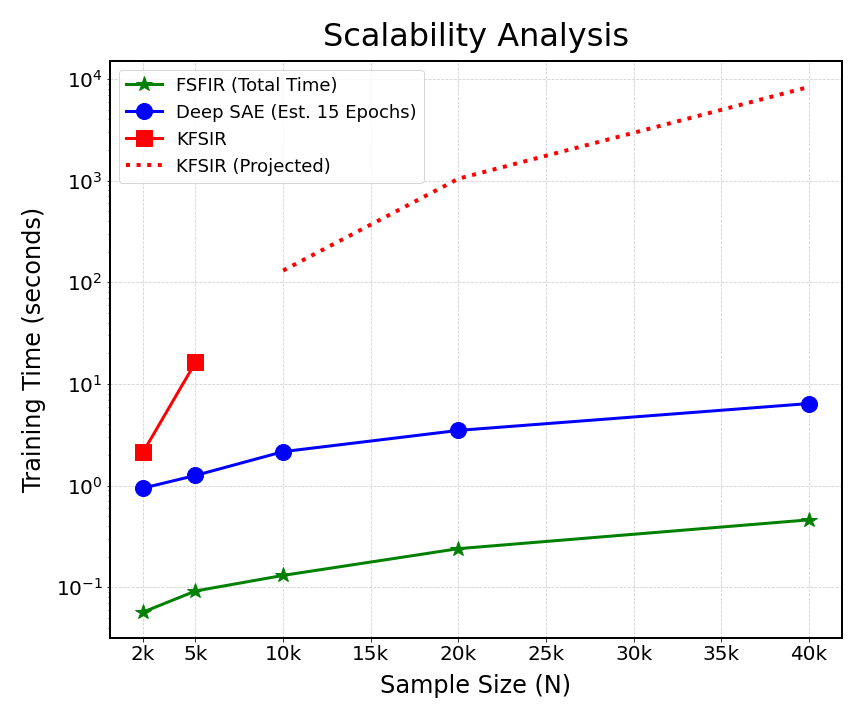}
    \caption{Efficiency Analysis. FSFIR dominates the Pareto frontier in Accuracy-vs-Time (Left) and exhibits near-linear practical scaling with highly optimized block-wise tensor contractions (Right).}
    \label{fig:scale_pareto}
\end{figure}

Executed on a standard GPU using PyTorch, performing this block-wise contraction for $N=40,000$ samples takes approximately \textbf{0.58} seconds and consumes negligible VRAM. As shown in Figure \ref{fig:scale_pareto}, FSFIR resides strictly in the optimal top-left quadrant of the Pareto frontier, fully unleashing its scalability to massive datasets without hardware constraints.

\subsection{Interpretability on Real-World Physics}
\label{sec:Exp_Interpretability}
Beyond accuracy and speed, FSFIR inherently produces a white-box representation. Applying FSFIR to the Bike Sharing dataset \citep{fanaee2014event} (predicting regression targets of daily rentals from continuous hourly temperature curves), we extract the learned functional weight $\beta(t)$.

\begin{figure}[t]
    \centering
    \includegraphics[width=0.45\textwidth]{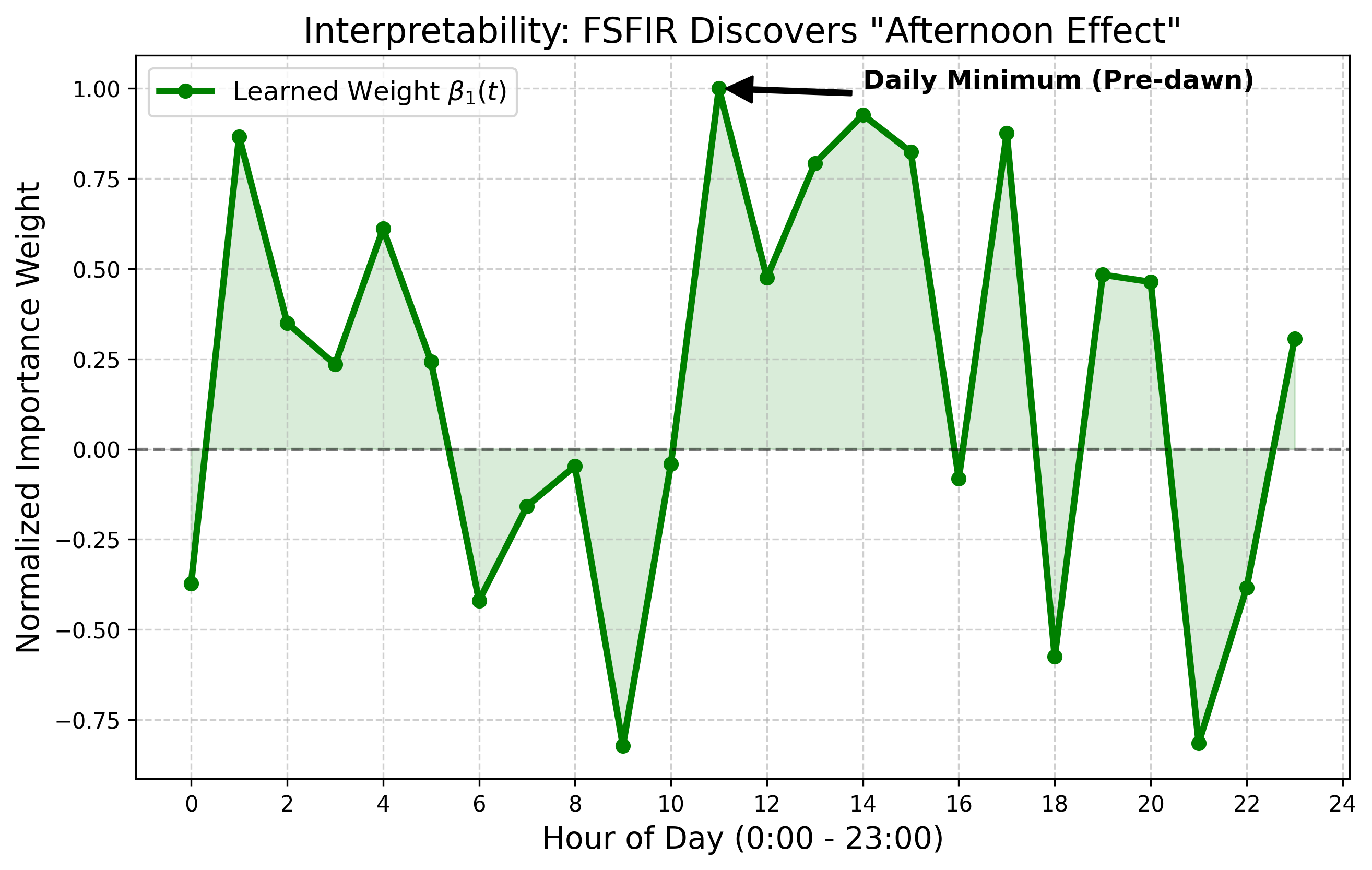}
    \caption{Interpretability on Bike Sharing. FSFIR automatically identifies physical thermal extremes (daily min/max temperatures) as critical drivers.}
    \label{fig:visual_bike}
\end{figure}

As illustrated in Figure \ref{fig:visual_bike}, rather than extracting uninterpretable latent vectors, FSFIR explicitly highlights the Daily Minimum (4:00 AM) and Maximum (2:00 PM) temperatures as the dominant predictive features. This precise alignment with physical/meteorological mechanisms demonstrates FSFIR's immense value in high-stakes domains requiring full functional transparency.

\section{Discussion}
\label{section, Discussion}

In this work, we introduced the Martingale Difference Divergence Operator (MDDO) and the FSFIR framework, bridging the gap between operator-theoretic statistics and modern representation learning. By formulating supervised dimensionality reduction as a closed-form spectral operator, FSFIR achieves the discriminative separation characteristic of supervised contrastive learning, but entirely bypasses negative sampling heuristics and gradient descent.

\paragraph{Broader Impact}
FSFIR offers a mathematically transparent and highly sample-efficient solution for continuous, high-dimensional signals. Its Safe-Fail stability and gradient-free nature make it particularly valuable in resource-constrained environments (e.g., edge devices) and mission-critical domains (e.g., healthcare physiological monitoring), where both rapid deployment and model interpretability are strictly required. Furthermore, its ability to act as an instantaneous projection head for frozen foundational models provides a green alternative to training massive deep classifiers.

\paragraph{Limitations and Future Directions}
While highly efficient, FSFIR relies on a fixed global spectral basis. For extremely complex geometries with highly localized frequency shifts, adaptive deep filters (e.g., Mamba operators) might offer greater expressivity. Future research could explore integrating MDDO as a differentiable spectral layer within end-to-end neural operators, combining the analytical stability of our operator with the adaptive flexibility of deep learning. Additionally, extending FSFIR to handle functional-to-functional mappings (e.g., sequence-to-sequence translation) remains a promising frontier.

\bibliography{references}

\clearpage
\appendix
\thispagestyle{empty}
\onecolumn
\mddotitle{MDDO: An Analytical Spectral Operator for Supervised Contrastive Representation Learning on Continuous Signals\\ (Supplementary Materials)}

\section{Additional Simulation Results of Section \hsecsynth}\label{sec:appendix:additional simulation}
This section contains the additional simulation results of Section \ref{sec:Exp_Synthetic} when $\varepsilon\sim \mc{N}(0,1)$.

We show the average $\mc D(\bs B;\bs{\wh B})$ with different $m$ or $\rho$ for three methods under $\mc M_1$ to $\mc M_3$ in Figure \ref{fig:error 3models,noise1},
where we mark minimal error in each model with red `$\times$'. The shaded areas represent the standard error associated with these estimates and all of them are less than $0.01$. For FSFIR, the minimal errors for $\mc M_1-\mc M_3$ are $0.08,0.02,0.01$ respectively.
For TFSIR, the minimal errors are $0.08,0.02,0.01$ and for regularized FSIR, the minimal errors are $0.13,0.06,0.01$. 

\begin{figure}[H]
	\centering
	\begin{minipage}{0.33\textwidth}
		\includegraphics[width=\textwidth]{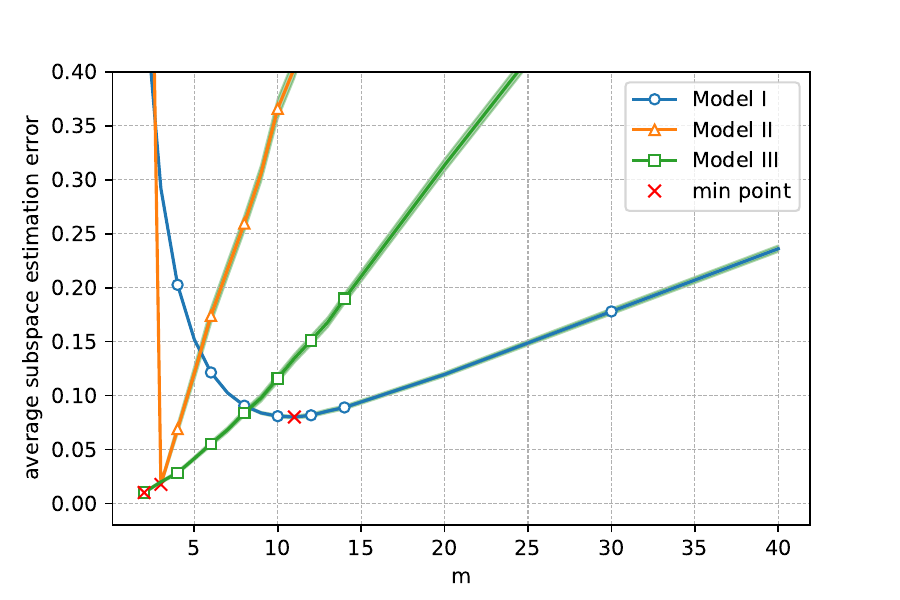}
	\end{minipage}\hfill
	\begin{minipage}{0.33\textwidth}
		\includegraphics[width=\textwidth]{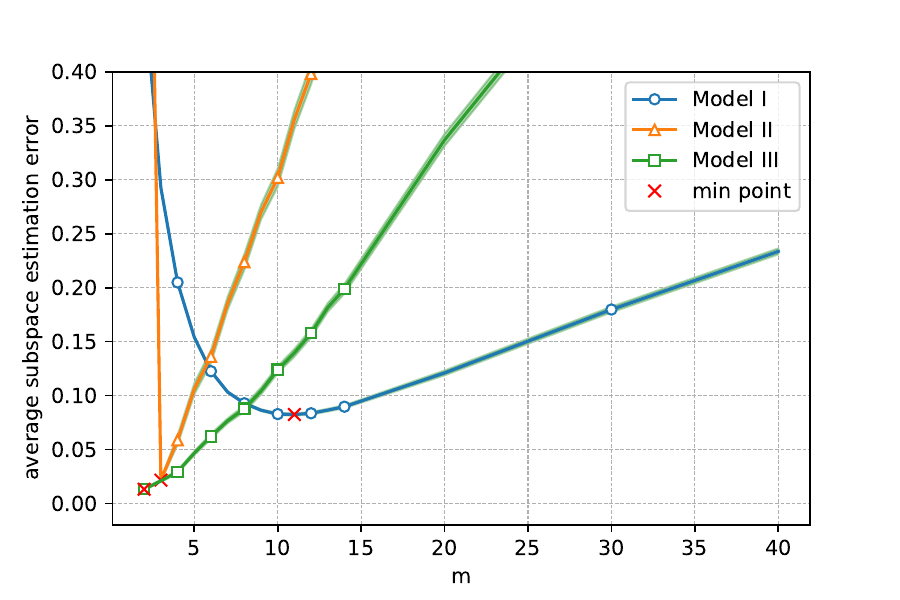}
	\end{minipage}
	\begin{minipage}{0.33\textwidth}
		\includegraphics[width=\textwidth]{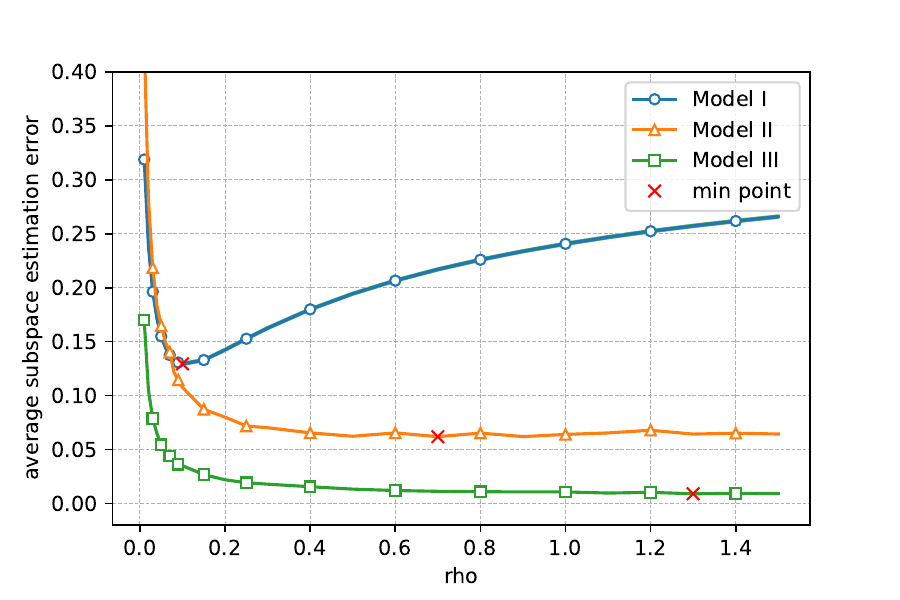}
	\end{minipage}
\caption{Average subspace estimation error of FSFIR (left), TFSIR (middle) and RFSIR (right) for various models. The standard errors are all below $0.01$. }
\label{fig:error 3models,noise1}
\end{figure}

Figure \ref{fig:error 3models,noise1} shows that FSFIR attains the best performance among all models. 
Moreover, FSFIR is easier to practice as it does not need a slice number $H$ in advance. 

\section{Proof of Lemma \hlemmaequiv}
\begin{proof}
For any ${\bs{\beta}}\in\mc H$, according to the definition of $G_{\bs s}$ (see Definition $\ref{def: MDDO}$), one has
\begin{align*}
\langle G_{\bs s},{\bs{\beta}}\rangle&=\int_{[0,1]} G_{\bs s}(t){\bs{\beta}}(t)~\mathrm{d}t=\int_{[0,1]}\mathrm{cov}\mleft(\bs{X}(t),\mathrm{e}^{\mi\langle \bs s,\Y\rangle}\mright){\bs{\beta}}(t)~\mathrm{d}t=\int_{[0,1]}\mathrm{cov}\mleft(\bs{X}(t){\bs{\beta}}(t),\mathrm{e}^{\mi \langle \bs s,\Y\rangle}\mright)~\mathrm{d}t.
\end{align*}
By Fubini theorem, under Assumption $\ref{as:joint distribution assumption}$, one can exchange the order of integration and covariance above and get that
\begin{align*}
 \langle G_{\bs s},{\bs{\beta}}\rangle&=\int_{[0,1]}\mathrm{cov}\mleft(\bs{X}(t){\bs{\beta}}(t),\mathrm{e}^{\mi \langle \bs s,\Y\rangle}\mright)~\mathrm{d}t=\mathrm{cov}\bigg(\int_{[0,1]}\bs{X}(t){\bs{\beta}}(t)~\mathrm{d}t,\mathrm{e}^{\mi \langle \bs s,\Y\rangle}\bigg)=\mathrm{cov}\mleft(\langle \bs{X},{\bs{\beta}}\rangle,\mathrm{e}^{\mi \langle \bs s ,\Y\rangle}\mright).
\end{align*}
Thus for any $\bs\alpha(t),{\bs{\beta}}(t)\in\mc H$, one can get
\begin{align*}
\big\langle \big(G_{\bs s}\otimes \overline{G}_{\bs s}\big)\bs\alpha,{\bs{\beta}}\big\rangle=\langle G_{\bs s},\bs\alpha\rangle\langle \overline{G}_{\bs s},{\bs{\beta}}\rangle=\mathrm{cov}\mleft(\langle \bs{X},\bs\alpha\rangle,\mathrm{e}^{\mi \langle \bs s,\Y\rangle}\mright)\mathrm{cov}\mleft(\langle \bs{X},{\bs{\beta}}\rangle,\mathrm{e}^{-\mi\langle \bs s,\Y\rangle}\mright)\\
=\mb{E}\mleft(\langle \bs{X},\bs\alpha\rangle\mathrm{e}^{\mi \langle \bs s,\Y\rangle}\mright)\mb{E}\mleft(\langle \bs{X},{\bs{\beta}}\rangle\mathrm{e}^{-\mi \langle \bs s,\Y\rangle}\mright)=\mb{E}\Big(\langle \bs{X},\bs\alpha\rangle\langle \bs{X}',{\bs{\beta}}\rangle\mathrm{e}^{\mi \langle \bs s,\Y-\Y'\rangle}\Big).
\end{align*}
Considering that $\mb{E}\big(\langle \bs{X},\alpha\rangle\langle \bs{X}',{\bs{\beta}}\rangle\big)=0$, one has
\begin{align*}
\big\langle \big(G_{\bs s}\otimes \overline{G}_{\bs s}\big)\bs\alpha,{\bs{\beta}}\big\rangle
&=- \mb{E}\Big(\langle \bs{X},\bs\alpha\rangle\langle \bs{X}',{\bs{\beta}}\rangle\big(1-\mr{e}^{\mi \langle \bs s,\Y-\Y'\rangle}\big)\Big)\\
&=- \mb{E}\Big(\langle \bs{X},\bs\alpha\rangle\langle \bs{X}',{\bs{\beta}}\rangle\big[1-\cos\big(\langle \bs s,\Y-\Y'\rangle\big)\big]\Big)
+\mi\mb{E}\Big(\langle \bs{X},\bs\alpha\rangle\langle \bs{X}',{\bs{\beta}}\rangle\big[\sin\big(\langle\bs s,\Y-\Y'\rangle\big)\big]\Big).
\end{align*}
It is easy to check that
\[\int_{\mb R^q}\frac{\sin \big(\langle\bs s,\Y-\Y'\rangle)\big)}{\|\bs s\|^{1+q}}~\mr{d}\bs s=\lim_{\varepsilon\to0^+}\int_{\bs s\in\mb{R}^q:\varepsilon\leqslant\|\bs s\|\leqslant \varepsilon^{-1}}\frac{\sin \big(\langle \bs s,\Y-\Y'\rangle\big)}{\|\bs s\|^{1+q}}~\mr{d}\bs s=0,\]
because the integrand is an odd function. 

By Lemma 1 in \cite{szekely2007measuring}, one can also get
\[\int_{\R^q}\frac{1-\cos\big(\langle \bs s,\Y-\Y'\rangle\big)}{\|\bs s\|^{1+q}}~\mr{d}\bs s=c_q\|\Y-\Y'\|.
\]
Combining above results with Definition $\ref{def: MDDO}$, one can obtain that 
\begin{align}\label{proof: lemma MDDO}
\langle\mathrm{MDDO}(\bs{X}|Y)\bs\alpha,{\bs{\beta}}\rangle=- \mb{E}\Big(\langle \bs{X},\bs\alpha\rangle\langle \bs{X}',{\bs{\beta}}\rangle\|\Y-\Y'\|\Big) .
\end{align}
Then by the arbitrariness of $\bs\alpha,{\bs{\beta}}\in\mc H$, the proof is completed. 
\end{proof}

\section{Proof of Theorem \htheoremcmi}

\begin{proof}[\textbf{Proof of Theorem \ref{theorem, MDDO and conditional mean independence}(ii)}]
For any test functions $\bs u, \bs v \in \mathcal{H}$, using the closed-form representation in Lemma \ref{lemma, equivalence of two def of MDDO} or \eqref{proof: lemma MDDO}, we have:
\begin{align*}
\langle\mathrm{MDDO}(T^*\bs{X}|\Y)\bs u, \bs{v}\rangle &= - \mathbb{E}\Big[ \langle T^*\bs{X}, \bs u \rangle \langle T^*\bs{X}', \bs{v} \rangle \|\Y-\Y'\| \Big]= - \mathbb{E}\Big[ \langle \bs{X}, T\bs u \rangle \langle \bs{X}', T\bs{v} \rangle \|\Y-\Y'\| \Big] \\
&= \langle\mathrm{MDDO}(\bs{X}|\Y)T\bs u, T\bs{v}\rangle= \langle T^*\mathrm{MDDO}(\bs{X}|\Y)T\bs u, \bs{v}\rangle.
\end{align*}
Since this equality holds for all $\bs u, \bs v \in \mathcal{H}$, we conclude that $\mathrm{MDDO}(T^*\bs{X}|\Y) = T^*\mathrm{MDDO}(\bs{X}|\Y)T$. 
\end{proof}

To prove (i), we proceed as follows. According to \eqref{proof: lemma MDDO}, one can get the following useful lemma.
\begin{lemma}\label{lemma, MDDO and FMDD}
Under Assumption $\ref{as:joint distribution assumption}$, for all ${\bs{\beta}}\in\mathcal H$, $\|{\bs{\beta}}\|=1$, we have
\begin{align*}
\langle \mathrm{MDDO}(\boldsymbol{X}|\Y)({\bs{\beta}}),{\bs{\beta}}\rangle &=- \mathbb E\Big[ \langle \boldsymbol{X},{\bs{\beta}}\rangle \langle \boldsymbol{X}',{\bs{\beta}}\rangle \|\Y-\Y'\|\Big]=- \mathbb E\Big[\big\langle\langle \boldsymbol{X},{\bs{\beta}}\rangle{\bs{\beta}},\langle \boldsymbol{X}',{\bs{\beta}}\rangle{\bs{\beta}}\big\rangle\|\Y-\Y'\|\Big].
\end{align*}
\end{lemma}
This conclusion links MDDO with functional martingale
difference divergence (FMDD \citep{lee2020testing}). 
Next we give the following two lemmas to finish the proof of Theorem $\ref{theorem, MDDO and conditional mean independence}$.
\begin{lemma}\label{lem: Txx=0tuiTx=0}If $T$ is a positive semi-definite operator on a Hilbert space $\wt{\mathcal{H}}$, then for all $x\in\wt{\mathcal{H}}$, one has $\langle Tx,x\rangle=0\Longleftrightarrow Tx=0$.
\end{lemma}
\begin{proof}
`$\Longleftarrow$': It is obvious.

`$\Longrightarrow$': It is easy to check that $f(a,b)=\langle Ta,b\rangle$ $(a,b\in\wt{\mc H})$ is a 
positive semi-definite Hermitian form. Thus, for any $y\in\wt{\mathcal{H}}$, one can use Cauchy inequality to get
\[|\langle Tx,y\rangle|^2\leqslant\langle Tx,x\rangle\langle Ty,y\rangle=0\Longrightarrow \langle Tx,y\rangle=0.\]
By the arbitrariness of $y\in\wt{\mc H}$, one has $Tx=0$.
\end{proof}

Our proof of Theorem $\ref{theorem, MDDO and conditional mean independence}$ is mainly inspired by the following property of
FMDD in \cite{lee2020testing}.
\begin{lemma}[Proposition 1 of \cite{lee2020testing}]\label{lem:prop1inlee}
If $\E[\|\X\|+\|\Y\|]<\infty$ and $\E[\|\bs X\|\|\Y\|]<\infty$, then we have
\[\E[\langle \X,\X'\rangle\|\Y-\Y'\|]=0\Longleftrightarrow \E[\X|\Y]=0\quad\text{almost surely},\]
where $(\X',\Y')$ is an i.i.d. copy of $(\X,\Y)$.
\end{lemma}
\begin{proof}[\textbf{Proof of Theorem \ref{theorem, MDDO and conditional mean independence}(i)}]
First, one has
\begin{align*}\mathrm{MDDO}(\boldsymbol{X}|\Y)=0 &\Longleftrightarrow \mathrm{MDDO}(\boldsymbol{X}|\Y)({\bs{\beta}})=0,\quad\forall{\bs{\beta}}\in\mb{S}_{\mathcal H};\\
\mathbb E[\boldsymbol{X}|\Y]=0~~\text{a.s.}&\Longleftrightarrow\langle\mb E[\boldsymbol{X}|\Y],{\bs{\beta}}\rangle{\bs{\beta}}=0~~\text{a.s.} \quad\forall{\bs{\beta}}\in\mb{S}_{\mathcal H},
\end{align*}
where $\mb{S}_\mc{H}=\{{\bs{\beta}}\in\mc H:\|{\bs{\beta}}\|=1\}$. Second, from Lemma $\ref{lem: Txx=0tuiTx=0}$, one knows that
\begin{align*}
\mathrm{MDDO}(\boldsymbol{X}|\Y)({\bs{\beta}})=0&\Longleftrightarrow\langle\mathrm{MDDO}(\boldsymbol{X}|\Y)({\bs{\beta}}),{\bs{\beta}}\rangle=0.
\end{align*}
 Then under Assumption $\ref{as:joint distribution assumption}$, by Lemma $\ref{lemma, MDDO and FMDD}$ and $\ref{lem:prop1inlee}$, one has
\begin{align*}
&\langle\mathrm{MDDO}(\boldsymbol{X}|\Y)({\bs{\beta}}),{\bs{\beta}}\rangle=0\Longleftrightarrow\mathbb E[\big\langle\langle \boldsymbol{X},{\bs{\beta}}\rangle{\bs{\beta}},\langle \boldsymbol{X}',{\bs{\beta}}\rangle{\bs{\beta}}\rangle\|\Y-\Y'\|]=0\\
&\qquad\qquad\qquad\qquad\qquad\Longleftrightarrow\mathbb E[\langle \bs X,{\bs{\beta}}\rangle{\bs{\beta}}|\Y]=\langle \mb E[\boldsymbol{X}|\Y],{\bs{\beta}}\rangle{\bs{\beta}}=0~~\text{a.s.}
\end{align*}
This finishes the proof of Theorem $\ref{theorem, MDDO and conditional mean independence}$.
\end{proof}


\section{Proof of Lemma \hlemmasegamma}
Recall the following fact in FSIR.
\begin{lemma}\label{lemma, direct result of linearity condition}
Under Assumption $\ref{as:Linearity condition and Coverage condition}~ \boldsymbol{\mathrm{{i)}}}$, we have $\mathcal S_{\mathbb E(\boldsymbol{X}|\Y)}\subseteq \Gamma \mc S_{\Y|\bs X}\subseteq \mc H$.
\end{lemma}
It is a trivial generalization of \cite[Theorem 2.1]{ferre2003functional} from univariate response to multivariate response.
\begin{proof}[\textbf{Proof of Lemma \ref{lemma: SE=GammaS}}]
First, we prove that $\mathcal{S}_{\mathbb{E}(\bs X|\Y)}^\perp\subseteq \mathrm{Im}\{\mathrm{var(\mb{E}(\bs X|\Y))}\}^\perp$. For any ${\bs{\beta}}\in\mathcal{S}_{\mathbb{E}(\bs X|\Y)}^\perp$, one has $\langle{\bs{\beta}},\mb{E}(\bs X|\Y)\rangle=0$ a.s. Then for any $\bs\alpha\in\mathcal{H}$, one can get
\begin{align*}
\langle{\bs{\beta}},\mathrm{var}(\mb{E}(\bs X|\Y))\bs\alpha\rangle&=\langle{\bs{\beta}},\mb E\lmi\mb{E}(\bs X|\Y)\otimes \mb{E}(\bs X|\Y)\rmi\bs\alpha\rangle\\
&=\mb E\big(\langle\mb{E}(\bs X|\Y),\bs\alpha\rangle\langle{\bs{\beta}},\mb{E}(\bs X|\Y)\rangle\big)=0,
\end{align*}
which means that ${\bs{\beta}}\in\mathrm{Im}\{\mathrm{var}(\mb{E}(\bs X|\Y))\}^\perp$. Moreover, one has
\begin{align*}\mathcal{S}_{\mathbb{E}(\bs X|\Y)}^\perp\subseteq \mathrm{Im}\{\mathrm{var}(\mb{E}(\bs X|\Y))\}^\perp
\Longrightarrow\overline{\mathcal{S}_{\mathbb{E}(\bs X|\Y)}}\supseteq\overline{\mathrm{Im}}\{\mathrm{var}(\mb{E}(\bs X|\Y))\}.
\end{align*}
Thus, $\overline{\mathrm{Im}}\{\mathrm{var}(\mb{E}(\bs X|\Y))\}\subseteq\overline{\mathcal{S}_{\mathbb{E}(\bs X|\Y)}}\subseteq\overline{\Gamma\mathcal{S}_{\Y|\bs X}}$ by Lemma $\ref{lemma, direct result of linearity condition}$. According to Assumption $\ref{as:Linearity condition and Coverage condition}$ \textbf{ii)}, one can get
\[\mathrm{dim}\left(\overline{\mathrm{Im}}\{\mathrm{var}(\mb{E}(\bs X|\Y))\}\right)=\mathrm{dim}\left(\overline{\mathcal{S}_{\mathbb{E}(\bs X|\Y)}}\right)=\mathrm{dim}(\overline{\Gamma\mathcal{S}_{\Y|\bs X}})=d.\]
One can complete the proof since finite dimension subspaces are closed.
\end{proof}

\section{Proof of Theorem \htheoremirs}
\begin{proof}
For convenience, we abbreviate $\mathrm{MDDO}(\boldsymbol{X}|\Y)$ to ${M}$. According to Theorem $\ref{theorem, MDDO and conditional mean independence}$ and Lemma $\ref{lem: Txx=0tuiTx=0}$, one can get
\begin{align*}{\bs{\beta}}\in\mathcal S_{\mb E(\boldsymbol{X}|\Y)}^\perp&\Longleftrightarrow\langle {\bs{\beta}},\mathbb E(\boldsymbol{X}|\Y)\rangle=0~~\text{a.s.}\Longleftrightarrow\mathbb E(\langle {\bs{\beta}},\boldsymbol{X}\rangle|\Y)=0~~\text{a.s.}\\
&\Longleftrightarrow\mathrm{MDDO}(\langle {\bs{\beta}},\boldsymbol{X}\rangle|\Y)=0\Longleftrightarrow\langle {M}{\bs{\beta}},{\bs{\beta}}\rangle=0\\
&\Longleftrightarrow{M}{\bs{\beta}}=0\Longleftrightarrow {\bs{\beta}}\in\mathrm{null}(M)=\overline{\mathrm{Im}}(M)^\perp,
\end{align*}
which means that $\mathcal S_{\mb E(\boldsymbol{X}|\Y)}^\perp=\overline{\mathrm{Im}}(M)^\perp$ and $\overline{\mathcal S_{\mb E(\boldsymbol{X}|\Y)}}=\overline{\mathrm{Im}}(M)$.
One can complete the proof since finite dimension subspaces are closed.
\end{proof}
\section{Proof of Lemma \hlemmatrunc}
Before proving Lemma $\ref{lemma, way of estimate truncate central subspace}$, we give the following lemma.
\begin{lemma}\label{lem: colPBP equal colPB operator}
Assume that $P$ is a bounded linear operator from a Hilbert space $\wt{\mc H}$ to itself and $B$ is a positive semi-definite operator from $\wt{\mc H}$ to itself. 
Then we have $\overline{\mathrm{Im}}(PBP^*)=\overline{\mathrm{Im}}(PB)$.
\end{lemma}
\begin{proof}
It suffices to show that $\mnull(BP^*)=\mnull(PBP^*)$. First, since $B$ is positive semi-definite, one has $\langle x,PBP^*x\rangle = \langle P^*x,BP^*x \rangle\geqslant 0~(\forall x\in\wt{\mc H})$. Thus $PBP^*$ is a positive semi-definite operator on $\wt{\H}$.
For any $y\in\wt{\H}$, we have 
\begin{align*}
PBP^*y=0\overset{(a)}{\Longleftrightarrow}\langle y,PBP^*y\rangle = \langle P^*y,BP^*y \rangle=0\overset{(b)}{\Longleftrightarrow} BP^*y=0. 
\end{align*}
where $(a)$ and $(b)$ come from Lemma $\ref{lem: Txx=0tuiTx=0}$.
Thus $\mnull(PBP^*)=\mnull(BP^*)$.
\end{proof}

\begin{proof}[\textbf{Proof of Lemma \ref{lemma, way of estimate truncate central subspace}}]
For convenience, we abbreviate $\mathrm{MDDO}(\boldsymbol{X}|\Y)$ and $\mathrm{MDDO}(\boldsymbol{X}^{(m)}|\Y)$ to ${M}$ and $M_m$ respectively. 

By Corollary $\ref{corollary, MDDO and central subspace}$, one can get $\Gamma\mathcal{S}_{\Y|\boldsymbol{X}}=\mathrm{Im}(M)$. Thus,
\begin{align}\label{eq: corollary, MDDO and central subspace}
\Pi_m\Gamma\mathcal{S}_{\Y|\boldsymbol{X}}=\Pi_m\mathrm{Im}(M)=\mathrm{Im}(\Pi_m M).
\end{align}
It is easy to check that
\begin{align}
\Gamma_m&:=\mathrm{var}(\bs X^{(m)})=\Pi_m\Gamma\Pi_m=\Pi_m\Gamma=\Gamma\Pi_m=\sum\limits_{i=1}^m\lambda_i\phi_i\otimes\phi_i.\label{eq: Gamma m def}
\end{align}
On the one hand, by the definition of $\mathcal{S}^{(m)}_{{\Y|\boldsymbol{X}}}$ and $\Gamma_m$ (see \eqref{eq: Gamma m def}), one can get
\begin{align}\label{eq:Pim Gamma S}
\Pi_m\Gamma\mathcal{S}_{\Y|\boldsymbol{X}}&=\Pi_m\Gamma\Pi_m\mathcal{S}_{\Y|\boldsymbol{X}}=(\Pi_m\Gamma)(\Pi_m\mathcal{S}_{\Y|\boldsymbol{X}})=\Gamma_m\mathcal{S}^{(m)}_{{\Y|\boldsymbol{X}}}.
\end{align}
On the other hand, one has $\overline{\mathrm{Im}}(\Pi_m M)=\overline{\mathrm{Im}}(\Pi_m M\Pi_m)$ by Lemma $\ref{lem: colPBP equal colPB operator}$. Since $\Pi_m M$ and $\Pi_m M\Pi_m$ are both of finite rank, one can further get
\begin{align*}
\mathrm{Im}(\Pi_m M)&=\overline{\mathrm{Im}}(\Pi_m M)=\overline{\mathrm{Im}}(\Pi_m M\Pi_m)=\mathrm{Im}(\Pi_mM\Pi_m).
\end{align*}
Then according to Theorem $\ref{theorem, MDDO and conditional mean independence}$(ii), one has
\begin{align}
\mathrm{Im}(\Pi_m M)=\mathrm{Im}(\Pi_mM\Pi_m)=\mathrm{Im}(M_m).\label{eq:Pim span M}
\end{align}
Combining \eqref{eq:Pim Gamma S}, \eqref{eq:Pim span M} with \eqref{eq: corollary, MDDO and central subspace}, one has $\Gamma_m\mathcal{S}^{(m)}_{{\Y|\boldsymbol{X}}}=\mathrm{Im}\{M_m\}$.
Finally, one can get \[\Gamma_m^\dagger\mathrm{Im}\{M_m\}=\Gamma_m^\dagger\Gamma_m\mathcal{S}^{(m)}_{{\Y|\boldsymbol{X}}}=\Pi_m\mathcal{S}^{(m)}_{{\Y|\boldsymbol{X}}}=\mathcal{S}^{(m)}_{{\Y|\boldsymbol{X}}}.\]
\end{proof}

\section{Wely Inequality for a Self-adjoint and Compact Operator}\label{ap:Wely inequality for self-adjoint and compact operators}
First, we show the following three results in standard functional analysis textbook.
\begin{lemma}[Spectral theorem]\label{thm: Spectral theorem}Let $\wt{\mathcal{H}}$ be a Hilbert space and $A:\wt{\mc{H}}\to\wt{\mc{H}}$ be a compact, self-adjoint operator. There is an at most countable orthonormal basis $\{\wt e_j\}_{j\in J}$ ($J=\{1,\cdots,n\}$ or $\mathbb{Z}_{\geqslant1}$) of $\wt{\mathcal{H}}$ and eigenvalues $\{\wt\lambda_j\}_{j\in J}$ with $|\wt\lambda_1|\geqslant|\wt\lambda_2|\geqslant\cdots\geqslant0$ converging to zero, such that
\begin{align*}
x=\sum_{j\in J}\langle x,\wt e_j\rangle \wt e_j;\qquad Ax=\sum_{j\in J}\wt\lambda_j\langle x,\wt e_j\rangle \wt e_j,\qquad x \in\wt{\mathcal{H}}.
\end{align*}
\end{lemma}

\begin{lemma}[Rayleigh's principle]\label{lem:Rayleigh operator}Let $A$ be a compact, self-adjoint operator. If $\{\wt e_j\}_{j\in J}$ and $\{\wt\lambda_j\}_{j\in J}$ are eigenvectors and eigenvalues define in Lemma $\ref{thm: Spectral theorem}$ respectively. Then
\[|\wt\lambda_1|=\mathop{\sup\limits_{\|u\|=1}}|\langle Au,u\rangle|;\qquad|\wt\lambda_n|=\mathop{\sup\limits_{\|u\|=1}}_{u\in\{\wt e_1,\cdots,\wt e_{n-1}\}^\perp}|\langle Au,u\rangle|~(n\geqslant 2).\]
\end{lemma}
\begin{lemma}[Minimax theorem]\label{lem:minimax operator}
Assume that $A$ is a positive semi-definite and compact operator with its eigenvalues $\{\wt\lambda_i\}$ ordered as $\wt\lambda_1\geqslant\dots\geqslant \wt\lambda_n\geqslant\dots\geqslant 0$, then
$$
\wt\lambda_n=\inf_{E_{n-1}}\sup_{x\in E_{n-1}^\perp,\|x\|=1}\langle Ax,x\rangle
$$
where $E_{n-1}$ with dimension $n-1$ is a closed linear subspace of $\wt{\mc H}$.
\end{lemma}
Then we give the Wely inequality for a self-adjoint and compact operator.
\begin{proposition}\label{prop: wely operator}
Let $M=N+R$ where $M$, $N$ and $R$ are three self-adjoint and compact operators defined on a Hilbert space $\wt{\mc H}$. Also, $M$ and $N$ are positive semi-definite with their respective eigenvalues $\{\mu_i\},\{\nu_i\}$ ordered as follows
\begin{align*}
M:\mu_1\geqslant\dots\geqslant \mu_n\geqslant\dots\geqslant 0;\qquad
N:\nu_1\geqslant\dots\geqslant \nu_n\geqslant\dots\geqslant 0,
\end{align*}
while $R$'s eigenvalues are $\{\rho_i\}$ ordered as follows:
\[R:|\rho_1|\geqslant\dots\geqslant |\rho_n|\geqslant\dots\geqslant 0.\]
Then the following inequalities hold: $|\mu_k-\nu_k|\leqslant|\rho_1|=\|R\| $, $k\geqslant1$.
\end{proposition}
\begin{proof}
From Lemma $\ref{lem:minimax operator}$, we have:
\[\mu_n=\inf_{E_{n-1}}\sup_{x\in E_{n-1}^\perp,\|x\|=1}\langle Mx,x\rangle;\qquad\nu_n=\inf_{E_{n-1}}\sup_{x\in E_{n-1}^\perp,\|x\|=1}\langle Nx,x\rangle,\]
where $E_{n-1}$ with dimension $n-1$ is a closed linear subspace of $\wt{\mc H}$.
By Lemma $\ref{lem:Rayleigh operator}$, we have:
$$
\sup_{\|u\|=1}|\langle Ru,u\rangle|=|\rho_1|.
$$
Since $\langle Mu,u\rangle=\langle Nu,u\rangle+\langle Ru,u\rangle$, for any $\|u\|=1$, we have:
$$
\langle Nu,u\rangle-|\rho_1|\leqslant\langle Mu,u\rangle \leqslant \langle Nu,u\rangle+|\rho_1|.
$$
Then for any given $n-1$ dimensional closed linear subspace of $\wt{\mc H}$, we conclude
\begin{equation}\label{eq:max ineq}
\sup_{u\in E_{n-1}^\perp,\|u\|=1}\langle Nu,u\rangle-|\rho_1|\leqslant\sup_{u\in E_{n-1}^\perp,\|u\|=1}\langle Mu,u\rangle\leqslant \sup_{u\in E_{n-1}^\perp,\|u\|=1}\langle Nu,u\rangle+|\rho_1|.
\end{equation}
Take the infimum with respective to $E_{n-1}$ in \eqref{eq:max ineq}, we have
\[\nu_n-|\rho_1|\leqslant\mu_n\leqslant \nu_n+|\rho_1|\]
by Lemma $\ref{lem:minimax operator}$.
\end{proof}
The next result is a direct corollary of Proposition $\ref{prop: wely operator}$.
\begin{corollary}\label{coro:wely ineq operator}
Let $M$ and $N$ be two self-adjoint, positive semi-definite and compact operators defined on a Hilbert space $\wt{\mc H}$ with their respective eigenvalues $\{\mu_i\},\{\nu_i\}$ ordered as follows
\begin{align*}
M:\mu_1\geqslant\dots\geqslant \mu_n\geqslant\dots\geqslant 0\quad\text{and}\quad
N:\nu_1\geqslant\dots\geqslant \nu_n\geqslant\dots\geqslant 0.
\end{align*}
Then the following inequalities hold: $|\mu_k-\nu_k|\leqslant\|M-N\| $, $ k\geqslant1$.
\end{corollary}

\section{Mathematical restatement of Algorithm \halgoslicing}
To facilitate subsequent proofs, this section formalizes some statements from the main text.
From now on, we abbreviate
$\mathrm{MDDO}(\boldsymbol{X}|\Y)$, $\mathrm{MDDO}(\boldsymbol{X}^{(m)}|\Y)$ to $M$ and $M_m$, respectively. 

We recall the estimation procedures. 
Define the estimator of $\Gamma_m$ and $M_m$ as follows:
\begin{align*}
\widehat{\Gamma}_m&:=\tfrac{1}{n}\textstyle\sum_{i=1}^n \bs X_i^{(m)}\otimes \bs X_i^{(m)};\\\wh M_m&:=- \tfrac{1}{n^2}\textstyle\sum_{j,k=1}^n\bs X_j^{(m)}\otimes \bs X_k^{(m)}\|\Y_j-\Y_k\|
\end{align*}
where $\bs X_i^{(m)}:=\Pi_m\bs X_i,i=1,...,n$. Note that standardizing $\X$ in Algorithm \ref{alg:slicing-free} is performed to ensure zero mean, but the theoretical part of this paper has already made the relevant assumptions.
Then the estimator of $\mc S_{\Y|\bs X}^{(m)}$ can be defined by
\begin{align*}
\wh{\mc S}_{\Y|\bs X}^{(m)}:=\widehat{\Gamma}_m^\dagger\mathrm{Im}^d\big\{\widehat M_m\big\}, 
\end{align*}
where $\mathrm{Im}^d\big\{\widehat M_m\big\}$ denotes the space spanned by the top $d$ eigenfunctions of $\wh M_m$.

An equivalent definition of $\wh{\mc S}_{Y|\bs X}^{(m)}$ can be derived as follows: define $\wh M_m^d$ by:
\begin{align}\label{wh M_m spectral decomposition}
\wh M_m=\textstyle\sum\limits_{i=1}^\infty\wh\mu_i\wh\gamma_i\otimes\wh\gamma_i\quad\text{and}\quad \wh M_m^d:=\textstyle\sum\limits_{i=1}^d\wh\mu_i\wh\gamma_i\otimes\wh\gamma_i,\end{align}
where $\lb\wh\mu_i\rb_{i\in\Z_{\geq1}}$ are eigenvalues with descending order of $\wh M_m$, and $\wh\gamma_i$'s are eigenfunctions of $\wh M_m$ associated with $\wh\mu_i$. 
Then
\begin{align}\label{def: estimator central subspace}
\wh{\mc S}_{\bs{Y}|\bs X}^{(m)}= \mathrm{Im}\big\{\widehat{\Gamma}_m^\dagger\widehat M_m^d\big\}.\end{align}

In addition, we state Assumption \ref{assumption: sub-Gaussian} under its complete mathematical formulation as follows:
\begin{assumption}[Complete version of Assumption \ref{assumption: sub-Gaussian}]\label{assumption: sub-Gaussian_Complete}

There exist two positive constants $\sigma_0$ and $\sigma_1$ such that\begin{align*}
\begin{split}
 \sup_m\max_{1\leqslant j\leqslant m}\mathbb E\left[\exp\big(2\sigma_0\langle \boldsymbol{X},\phi_j\rangle^2\big)\right]\leqslant \sigma_1&\\\quad\text{and} \quad \mathbb E\left[\exp\big(2\sigma_0\|\Y\|^2\big)\right]\leqslant \sigma_1&. 
 \end{split}\end{align*}

There exist positive constants $\alpha_1$, $\alpha_2$, $\wt C$ and $\wt C'$ such that for all $i\in[d]$, $j\in\mb Z_{+}$, we have
\begin{align*}
\alpha_2>1/2,\quad\lambda_j\geqslant \wt Cj^{- \alpha_1}\quad\text{and}\quad |b_{ij}|\leqslant \wt C'j^{- \alpha_2}, 
\end{align*}
 where $b_{ij}:=\langle  \bs{\beta}_i,\phi_j\rangle$ for $\bs\beta_i$ defined in \eqref{def: central subspace}.
\end{assumption}

\begin{proposition}\label{prop:bound hatMmd Mm}
Under Assumptions $\ref{as:joint distribution assumption}$ and $\ref{assumption: sub-Gaussian}$, for all $\gamma\in(0,1/2)$, there exist positive constants $D_0=D_0(\gamma,\sigma_0,\sigma_1)$, $D_1=D_1(\sigma_1)$, $D_2=D_2(\sigma_0,\sigma_1)$ and $n_0=n_0(\gamma,\sigma_0,\sigma_1)$ such that for all $n\geqslant n_0$ and
$$C\in \l D_0n^{{2\gamma}/{5}}-\ln\mpt{D_1m^2n},D_2 n^{{1}/{5}}-\ln\mpt{D_1m^2n} \rmi,$$ 
we have
\begin{equation*}
\mathbb{P}\mbk{\|\wh M_m^d- M_m\|\geqslant\l \tfrac{C+\ln( D_1m^2n)}{D_2}\r^{\frac52}\tfrac{24m}{\sqrt n}}\leqslant \exp({- C}).
\end{equation*}
\end{proposition}

\section{Proof of Proposition \hpropbound}
Before proving Proposition $\ref{prop:bound hatMmd Mm}$, we give the following conclusion, whose proof is deferred to the end of this section.
\begin{proposition}\label{proposition, concentration of MDDO}
Under Assumptions $\ref{as:joint distribution assumption}$ and $\ref{assumption: sub-Gaussian}$, for all $\gamma\in(0,1/2)$, there exist positive constants $D_0=D_0(\gamma,\sigma_0,\sigma_1)$, $D_1=D_1(\sigma_1)$, $D_2=D_2(\sigma_0,\sigma_1)$ and $n_0=n_0(\gamma,\sigma_0,\sigma_1)$ such that for all $n\geqslant n_0$ and
\[C\in \l D_0n^{\frac{2\gamma}{5}}-\ln\l D_1m^2n \r,D_2 n^{\frac{1}{5}}-\ln\l D_1m^2n \r \rmi,\]
we have
\begin{equation*}
\mathbb{P}\l\left\|\wh M_m- M_m\right\| <\l \frac{C+\ln( D_1m^2n)}{D_2}\r^{\frac52}\frac{12m}{\sqrt n}\r\geqslant 1-\exp(- C).
\end{equation*}
\end{proposition}
\begin{proof}[\textbf{Proof of Proposition \ref{prop:bound hatMmd Mm}}]

Using Corollary $\ref{coro:wely ineq operator}$, one can get
$
\lambda_i\l\wh M_m\r\leqslant \lno\wh M_m-M_m\rno +\lambda_i\l M_m\r
$. 
Since $\rank(M_m)=d$, one can get $\lambda_i(M_m)=0,~i\geqslant d+1$. Thus by Proposition $\ref{proposition, concentration of MDDO}$, one has
\begin{align}\label{eq:lambdai hat Mm upper bound}
\mathbb{P}\l\lambda_{d+1}(\wh M_m)<\l \frac{C+\ln\l D_1m^2n\r}{D_2}\r^{\frac52}\frac{12m}{\sqrt n}\r\geqslant 1-\exp(- C)\qquad(i\geq d+1). 
\end{align}
Notice that 
\begin{align*}\lno\wh M_m^d- M_m\rno &\leqslant\lno M_m-\wh M_m\rno +\lno\wh M_m-\wh M_m^d\rno ;\\
\lno\wh M_m-\wh M_m^d\rno &=\left\|\sum_{i=d+1}^\infty\wh\mu_i\wh\gamma_i\otimes \wh\gamma_i\right\| =\widehat{\lambda}_{d+1}=\lambda_{d+1}(\widehat{M}_m)
\end{align*}
by \eqref{wh M_m spectral decomposition}.
Then combing Proposition $\ref{proposition, concentration of MDDO}$ with \eqref{eq:lambdai hat Mm upper bound} can complete the proof.

\end{proof}

\begin{proof}[\textbf{Proof of Proposition \ref{proposition, concentration of MDDO}}]
Note that $\boldsymbol{X}^{(m)}=\sum\limits_{j=1}^m\langle \boldsymbol{X},\phi_j\rangle\phi_j$, then a simple calculation leads to
\begin{align*}
M_m&=-\sum_{i,j=1}^m\mathbb E\big[\langle \boldsymbol{X},\phi_i\rangle\langle \boldsymbol{X}',\phi_{j}\rangle\|\Y-\Y'\|\big]\phi_i\otimes\phi_j;\\
\widehat{M}_m&=-\sum_{i,j=1}^m\frac1{n^2}\sum_{k,\ell=1}^n\langle \boldsymbol{X}_k,\phi_i\rangle\langle \boldsymbol{X}_\ell,\phi_j\rangle\|\Y_k-\Y_\ell\|\phi_i\otimes\phi_j.
\end{align*}

For a operator $\Gamma'$ that can be expanded as $\Gamma':=\sum\limits_{i,j=1}^\infty a_{ij}\phi_i\otimes\phi_{j}$, let us define its maximal norm as $\|\Gamma'\|_{\mathrm{max}}=\sup\limits_{i,j}|a_{ij}|$.

\begin{lemma}\cite[Theorem 1]{mai2021slicing}\label{lemma, concentration of MDDOnm}
Under Assumptions $\ref{as:joint distribution assumption}$ and $\ref{assumption: sub-Gaussian}$, for all
$\gamma\in(0,1/2)$, there exist positive
constants $C_0=C_0(\gamma,\sigma_0,\sigma_1)$, $C_1=C_1(\sigma_1)$, $C_2 = C_2(\sigma_0;\sigma_1)$ and $n_0 = n_0(\gamma,\sigma_0,\sigma_1)$
such that for all $n\geqslant n_0$ and $\varepsilon\in(C_0 n^{-(1/2-\gamma)},1]$, we have
\begin{equation*}
\mathbb{P}\l\lno \widehat{M}_m-M_m\rno_{\max}>12\varepsilon\r\leqslant C_1 m^2n\exp\l- C_2\l\varepsilon^2 n\r^{1/5}\r.
\end{equation*}
\end{lemma}
\noindent Since $\lno\widehat{M}_m-M_m\rno \leqslant m\lno\widehat{M}_m-M_m\rno_{\mathrm{max}}$, one has
\begin{equation*}
\mathbb{P}\l\lno\widehat{M}_m-M_m\rno >12m\varepsilon\r\leqslant C_1 m^2n\exp\l-C_2\l\varepsilon^2 n\r^{1/5}\r.
\end{equation*}
Let $C=C_2\l\ve^2n\r^{1/5}-\ln\l C_1m^2n\r$ satisfying 
\begin{align*}
C\in\l C_2C_0^{2/5}n^{2\gamma/5}-\ln\l C_1m^2n\r,C_2n^{1/5}-\ln\l C_1m^2n\r\rmi,
\end{align*}
then one has
\begin{equation*}
\mathbb{P}\l\lno\widehat{M}_m-M_m\rno \leqslant\l \frac{C+\ln\l C_1m^2n\r}{C_2}\r^{\frac52}\frac{12m}{\sqrt{n}}\r>1- \exp(- C).
\end{equation*}
Then in order to complete the proof, one only need to choose $D_0$, $D_1$ and $D_2$ to be $C_2C_0^{2/5}$, $C_1$ and $C_2$ respectively. 
\end{proof}

\section{Properties of Sub-Gaussian Random Vectors}
We first review the definition of sub-Gaussian random variables.
\begin{definition}[Sub-Gaussian random variable and its upper-exponentially bounded constant]\label{def:sub gaussian variable}
A random variable $X$ is called a sub-Gaussian random variable if $X$ satisfies one of the following equivalent properties:
\begin{itemize}
 \item[1).] Tails. $\P(|X|>t)\leqslant \exp(1-t^{2}/K^{2}_{1})$ for any $t>0$;
 \item[2).] Moments. $\E[|X|^{p}]^{1/p}\leqslant K_{2}\sqrt{p}$ for any $p\geqslant 1$;
 \item[3).]Super-exponential moment: $\E[\exp(X^{2}/K^{2}_{3})]\leqslant \mr{e}$.

\noindent Moreover, if $\E[X]=0$, then the properties $1)-3)$ are also equivalent to the following one:
\item[4).] Moment generating function: $\E[\exp(tX)]\leqslant \exp(t^{2}K^{2}_{4})$ for all $t\in\R$.
\end{itemize}
Here $K_1$, $K_2$, $K_3$ and $K_4$ are four constants.
$K$ is called an upper-exponentially bounded constant of $X$ if 
$K\geqslant \max\{K_{1},K_{2},K_{3},K_{4}\}$.
\end{definition}
\begin{definition}[Sub-Gaussian random vector and its upper-exponentially bounded constant]\label{def,sub-Gaussian random vector,upper-exponentially bounded constant}
 ${X}\in\R^m$ is called a sub-Gaussian random vector if for all $x\in\R^m$, one-dimensional marginal $\langle{X},x\rangle$ is sub-Gaussian random variable. $K$ is called an upper-exponentially bounded constant of $X$ if $K$ satisfies:
 \begin{align*}
K\geqslant \sup_{x\in\mb{S}^{m-1}}K(\langle X,x\rangle) 
 \end{align*}
 where $K(\langle X,x\rangle)$ denotes an upper-exponentially bounded constant of $\langle X,x\rangle$.
Moreover, $K$ is called a uniform (about $m$) upper-exponentially bounded constant of $X$ if $K$ satisfies:
 \begin{align*}
K\geqslant \sup_m\sup_{x\in\mb{S}^{m-1}}K\l \langle X,x\rangle\r.
 \end{align*}
Furthermore, $X$ is called a uniform (about $m$) sun-Gaussian random vector.
 \end{definition}
The following is an application of sub-Gaussian random vectors.
\begin{lemma}[\citealt{vershynin2010introduction}]\label{lem:esgrm}
 Let $\M=[\bs m_1~\cdots~\bs m_n]$ be an $m\times n$ matrix ($n>m$) whose columns $\m_{i}$ are 
 independent centered sub-Gaussian random vectors with 
 covariance matrix $\mathbf{I}_{m}$. Let $\sigma^{+}_{\min}(\M)$ and $\sigma_{\max}(\M)$ be the infimum and supremum of positive singular values of $\M$ respectively. Then, for any $t>0$, with probability at least $1-2\exp(- C^{\prime}t^{2})$, we have
 \begin{equation*}
 \sqrt{n}-C_0\sqrt{m}-t\leqslant \sigma^{+}_{\min}(\M)\leqslant \sigma_{\max}(\M)\leqslant \sqrt{n}+C_0\sqrt{m}+t
 \end{equation*}
 where $C'$ and $C_0$ are two positive constants depending only on $K(\bs m_1)$:
 the upper-exponentially bounded constant of $\bs m_1$.
\end{lemma}
\noindent Let $t=\sqrt m$, then one can easily get
\begin{align}\label{equation, min max eval}
\begin{split}
\lambda_{\max}\left(\frac1n \M\M^\top\right)\leqslant \left(1+\frac{(C_0+1)\sqrt m}{\sqrt n}\right)^2;\\
\lambda_{\min}^+\left(\frac1n \M\M^\top\right)\geqslant \left(1-\frac{(C_0+1)\sqrt m}{\sqrt n}\right)^2, 
\end{split}
\end{align}
with probability at least $1-2\exp(- C'm)$ where $\lambda^{+}_{\min}(\cdot)$ and $\lambda_{\max}(\cdot)$ stands for the infimum and supremum of the positive spectrum respectively.

\begin{lemma}\label{lemma, estiamtion error of inverse sample cov}
Assume that $\x_1,\x_2,...,\x_n$ are $n$ i.i.d. samples from an $m$-dimensional centered sub-Gaussian vector with an invertible covariance matrix $\Sigma$. Let $\wh\Sigma:=\frac1n\sum_i \x_i\x_i^\top$.
Then there exists a positive constant $n_1'=n_1'(K(\bs m_1),c_1)$ ($c_1$ is defined as $m=n^{c_1},\quad c_1\in(0,1)$), such that when $n\geqslant n_1'$, we have
\begin{align*}
\lno\wh{\Sigma}-\Sigma\rno&\leqslant (C_0+2)^2\lambda_{\max}(\Sigma)\sqrt{\frac mn}~~\text{and}~~ \lno\wh{\Sigma}^{-1}-\Sigma^{-1}\rno\leqslant \frac{4(C_0+2)^2}{\lambda_{\min}(\Sigma)}\sqrt{\frac mn},
 \end{align*}
 with probability at least $1-2\exp(- C'm)$, where $C'$ is defined in Lemma $\ref{lem:esgrm}$.
\end{lemma}
\begin{proof}
Let $\x_i=\Sigma^{\frac12}\m_i$ and $\bs{M}=[\bs m_1~\cdots~\bs m_n]$ where $\m_i$ is a centered sub-Gaussian random vector with covariance $\mathbf I_m$. Then one has 
\begin{align*}
\lno\wh\Sigma-\Sigma\rno&\leqslant\lno\Sigma^{\frac12}\rno\cdot\left\|\frac1n \M\M^\top-\mathbf I\right\|\cdot\lno\Sigma^{\frac12}\rno\\
&= \lambda_{\max}(\Sigma)\cdot\left[\lambda_{\max}\left(\frac1n \M\M^\top\right)-1\right]
\end{align*}
and 
\begin{align*}
\lno\wh{\Sigma}^{- 1}-\Sigma^{- 1}\rno
&\leqslant \lno\Sigma^{-\frac12}\rno\cdot\left\|\frac1n \M\M^\top-\mathbf I\right\|\cdot\lno\l\frac1n \M\M^\top\r^{-1}\rno\cdot\lno\Sigma^{-\frac12}\rno\\
&=\frac{1}{\lambda_{\min}(\Sigma)}\left[\lambda_{\max}\left(\frac1n \M\M^\top\right)-1\right]\cdot\lambda_{\min}\left(\frac1n \M\M^\top\right)^{-1}.
\end{align*}
By \eqref{equation, min max eval}, it is easy to check that
\begin{align*}&\lambda_{\max}\left(\frac1n \M\M^\top\right)-1\leqslant\left(1+\frac{(C_0+1)\sqrt m}{\sqrt n}\right)^2-1\leqslant\frac{(C_0+2)^2\sqrt m}{\sqrt n};\\
&\lambda_{\min}\left(\frac1n \M\M^\top\right)\geqslant \left(1-\frac{(C_0+1)\sqrt m}{\sqrt n}\right)^2\geqslant \frac14~\text{for}~n\geqslant [2(C_0+1)]^{\frac2{1-c_1}},
\end{align*}
with probability at least $1-2\exp(- C'm)$. Thus the proof is completed by choosing $n_1'(C_0,c_1):=[2(C_0+1)]^{\frac{2}{1-c_1}}$. 
\end{proof}

\begin{proposition}
\label{prop:concentration Gammam dag Mmd}
 Suppose that Assumptions $\ref{as:joint distribution assumption}$ to $\ref{assumption: rate-type condition}$ hold, then $\forall \gamma\in(0,1/2)$, we have
\begin{equation*}
\begin{aligned}
 \lno\widehat\Gamma_m^\dagger \widehat M_m^d-\Gamma_m^\dagger M_m\rno =O_{\mb{P}}({m^{\alpha_1+1}}/{n^{1/2-\gamma}}).
\end{aligned}
\end{equation*}
\end{proposition}

\section{Proof of Proposition \hpropconc}\label{ap:concentration inequality}
We first give the following lemma whose proof is deferred to the end of this section.
\begin{lemma}\label{lem:PimTPimtoT}If $T$ is of finite rank, then we have $\lim\limits_{m\to \infty}\|\Pi_m T\Pi_m-T\| =0$.
\end{lemma}
A direct corollary of this lemma is as follows.
\begin{corollary}\label{lemma, M go to Mm}
Under Assumptions $\ref{as:joint distribution assumption}$ and $\ref{as:Linearity condition and Coverage condition}$, we have $\lim\limits_{m\to\infty}\|M-M_m\| =0$.
\end{corollary}
\noindent We denote by $m_M(\varepsilon)$ the minimal integer $m_M$ satisfying $\|M-M_m\| \leqslant \varepsilon$ for all $m\geqslant m_M$.

Proposition $\ref{prop:concentration Gammam dag Mmd}$ is a direct corollary of the following Proposition.
\begin{proposition}
\label{prop:bound of finite estimate}
 Suppose that Assumptions $\ref{as:joint distribution assumption}$ to $\ref{assumption: rate-type condition}$ hold, then $\forall \gamma\in(0,1/2)$, there exist positive constants
 \begin{align*}
 n_1=n_1(\gamma,\sigma_0,\sigma_1,\bs K,m_M(1),c_1),\quad D_3=D_3(\|M\| ,\wt C,\bs K) 
 \end{align*}
and $C'=C'(\bs K)$
, such that when $n\geqslant n_1$, we have
\begin{equation*}
\begin{aligned}
\mb P\l \lno\widehat\Gamma_m^\dagger \widehat M_m^d-\Gamma_m^\dagger M_m\rno \leqslant \left[\frac{C+\ln(D_1m^2n)}{D_2}\right]^{\frac52}\frac{24m^{\alpha_1+1}}{\wt C\sqrt n}+D_3\frac{m^{(2\alpha_1+1)/2}}{n^{1/2}} \r&\\
\geqslant 1-\exp(- C)-2\exp(- C'm).&
\end{aligned}
\end{equation*}
Here, $D_1,D_2$ and $C$ are defined in Proposition $\ref{prop:bound hatMmd Mm}$ and $\bs K$ is the uniform upper-exponentially bounded constant of $(\sqrt{\lambda_1}w_1,\dots,\sqrt{\lambda_m}w_m)$. 
\end{proposition}
\begin{proof}
By triangle inequality, one has
\begin{align*}
\lno\widehat{\Gamma}_m^\dagger \widehat M_m^d-\Gamma_m^\dagger M_m\rno 
&=\lno\widehat\Gamma_m^\dagger \widehat M_m^d-\wh\Gamma_m^\dagger M_m+\wh\Gamma_m^\dagger M_m-\Gamma_m^\dagger M_m\rno 
\\&\leqslant\lno\Gamma_m^\dagger\rno \cdot \lno\widehat M_m^d-M_m\rno +\lno\widehat\Gamma_m^\dagger-\Gamma_m^\dagger\rno \cdot \lno M_m\rno .
\end{align*}
Thus one can bound $\lno\Gamma_m^{\dag}M_m-\widehat\Gamma_m^{\dag}\widehat M_m^d\rno $ by bound $\lno\Gamma_m^\dagger\rno $, $\lno\widehat\Gamma_m^\dagger-\Gamma_m^\dagger\rno $, $\lno\widehat M_m^d-M_m\rno $ and $\lno M_m\rno $ respectively.
\begin{itemize}
 \item\textbf{Bound of $\lno\Gamma_m^\dagger\rno $}: By Assumption $\ref{assumption: rate-type condition}$, one has 
\begin{align}\label{eq:bound Gammam dagger}
\lambda_j\geqslant \wt C j^{-\alpha_1}\Rightarrow\lno\Gamma_m^\dagger\rno =\lambda_m^{-1}\leqslant \wt{C}^{-1} m^{\alpha_1}. 
\end{align} 
 \item\textbf{Bound of $\lno\widehat\Gamma_m^\dagger-\Gamma_m^\dagger\rno $}:
 Let us define $\mc H_m:=\mathrm{span}\{\phi_1,\dots,\phi_m\}$ where $\{\phi_i\}$ is introduced in Equation $\eqref{eq:X expansion}$. It is easy to check that
$\lno\widehat\Gamma_m^\dagger-\Gamma_m^\dagger\rno =\lno(\widehat\Gamma_m^\dagger-\Gamma_m^\dagger)|_{\mc H_m}\rno $ since $\l\widehat\Gamma_m^\dagger-\Gamma_m^\dagger\r{\bs{\beta}}=0$ for any ${\bs{\beta}}\in\mc{H}_m^\perp$. 
Because $\l\widehat\Gamma_m^\dagger-\Gamma_m^\dagger\r|_{\mc H_m}$ can be represented by matrix $\widehat{\Sigma}^{-1}-\Sigma^{-1}$ defined in Lemma $\ref{lemma, estiamtion error of inverse sample cov}$ under orthonormal basis $\{\phi_i\}_{i=1}^m$, one can get $\lno\widehat\Gamma_m^\dagger-\Gamma_m^\dagger\rno =\|\widehat{\Sigma}^{-1}-\Sigma^{-1}\|$.
Similarly, one can also get $\lno\Gamma_m^\dagger\rno =\lno\Sigma^{-1}\rno=\lambda_{\min}^{-1}(\Sigma)$. Thus, by Lemma $\ref{lemma, estiamtion error of inverse sample cov}$ one has
\[\mb P\l\lno\widehat\Gamma_m^\dagger-\Gamma_m^\dagger\rno \leqslant {4(C_0+2)^2}\lno\Gamma^{\dag}_m\rno \sqrt{\frac mn}\r\geqslant 1-2\exp(- C'm)\]
for sufficiently large $n\geqslant n_1'(\bs K,c_1)$
. Combing with $\lno\Gamma_m^\dagger\rno \leqslant \wt{C}^{-1} m^{\alpha_1}$, one can get
\begin{equation}\label{eq: distance hat gamma m dagger hat gamma m dagger}
\mb P\l\lno\widehat\Gamma_m^\dagger-\Gamma_m^\dagger\rno \leqslant \frac{4(C_0+2)^2m^{(2\alpha_1+1)/2}}{\wt Cn^{1/2}}\r\geqslant 1-2\exp(- C'm)
\end{equation}
for sufficiently large $n\geqslant n_1'(\bs K,c_1)$.
 \item\textbf{Bound of $\lno\widehat M_m^d-M_m\rno $}:
 See Proposition $\ref{prop:bound hatMmd Mm}$.
 \item \textbf{Bound of $\lno M_m\rno $}: By Corollary $\ref{lemma, M go to Mm}$, $\|M-M_m\| \leqslant 1$ for sufficiently large $m\geqslant m_M(1)$. Then by triangle inequality, one can get
$\|M_m\| -\|M\| \leqslant \|M-M_m\| \leqslant 1$. Hence,
\begin{align}\label{eq:Mm leq M C}
\|M_m\| \leqslant \|M\| +1.
\end{align}
\end{itemize}
Combing Equations \eqref{eq:bound Gammam dagger}, \eqref{eq: distance hat gamma m dagger hat gamma m dagger}, Proposition $\ref{prop:bound hatMmd Mm}$ with \eqref{eq:Mm leq M C}, one can choose $D_3$ and $n_1$ to be ${4(C_0+2)^2(\|M\| +1)}/{\wt C}$ and $\max\{n_0,n_1'(\bs K,c_1),m_M(1)^{1/c_1}\}$ respectively to
complete the proof where $n_0$ is defined in Proposition $\ref{prop:bound hatMmd Mm}$.
\end{proof}

\begin{proof}[\textbf{Proof of Lemma \ref{lem:PimTPimtoT}}]By the triangle inequality and compatibility of operator norm, one has
\begin{align*}
\|\Pi_m T\Pi_m-T\| &\leqslant\|\Pi_mT\Pi_m-\Pi_mT\| +\|\Pi_mT-T\| \\
&\leqslant\|(\Pi_m-I)T^*\| +\|(\Pi_m-I)T\| 
\end{align*}
where $I=\sum\limits_{i=1}^\infty\phi_i\otimes\phi_i$ for $\{\phi_i\}_{i\in\mb{Z}_{\geqslant 1}}$ defined in \eqref{eq:X expansion} being an orthonormal basis of $\mc H$. 

Since $T$ is of finite rank, let us assume that $\{e_i\}_{i=1}^k$ is an orthonormal basis of $\mathrm{Im}(T)$ where $k=\mr{rank}(T)$. For any ${\bs{\beta}}\in\mathcal{H}$ such that $\|{\bs{\beta}}\|=1$, one has $\|T{\bs{\beta}}\|\leqslant\|T\| \|{\bs{\beta}}\|=\|T\| $, so one can assume that $T{\bs{\beta}}\in\mathrm{Im}(T)$ admits the following expansion under basis $\{e_i\}_{i=1}^k$:
\[T{\bs{\beta}}=\sum_{i=1}^k b_ie_i,\quad \sum_{i=1}^k b^2_i\leqslant\|T\| ^2<\infty.\]
Thus
\[\|(I-\Pi_m)T{\bs{\beta}}\|=\left\|\sum_{i=1}^k(I-\Pi_m) b_ie_i\right\|\leqslant\sum_{i=1}^k |b_i|\cdot\|(I-\Pi_m) e_i\|.\]
Clearly, $\|(\Pi_m-I)\alpha\|~(\forall\alpha\in\H)$ tends to $0$ as $m\to\infty$ since 
\[(I-\Pi_m)\alpha=\left(\sum_{i={m+1}}^\infty\phi_i\otimes\phi_i\right)\left(\sum\limits_{i=1}^\infty c_i\phi_i\right)=\sum_{i=m+1}^\infty c_i\phi_i\xrightarrow{m\to\infty} 0\]
where we have assumed that $\alpha=\sum\limits_{i=1}^\infty c_i\phi_i$ .

Thus $\forall\varepsilon>0$, there exists some $N_i>0$ such that $\forall m> N_i$ one has $\|(\Pi_m-I)e_i\|<\varepsilon$, $(\forall i=1,...,k)$. Let $N=\max\{N_1,\cdots,N_k\}$, then $\forall m>N$ one has
\[\|(I-\Pi_m)T{\bs{\beta}}\|\leqslant\sum_{i=1}^k |b_i|\cdot\|(I-\Pi_m) e_i\|\leqslant\sum_{i=1}^k |b_i|\varepsilon\leqslant k\varepsilon\|T\| ,\]
which means that $\forall m>N$, one has
\begin{align*}
\|(\Pi_m-I)T\| &=\sup_{\|{\bs{\beta}}\|=1}\|(\Pi_m-I)T{\bs{\beta}}\|\leqslant k\varepsilon\|T\| . 
\end{align*}
Thus $\lim\limits_{m\to\infty}\|(\Pi_m-I)T\| =0$. 

Similarly, one can also get $\lim\limits_{m\to\infty}\|(\Pi_m-I)T^*\| =0$. Then the proof of Lemma $\ref{lem:PimTPimtoT}$ is completed.
\end{proof}

\section{Sin Theta Theorem}\label{ap:Sin Theta theorem}
\subsection{Sin Theta Theorem for Self-adjoint Operators}
\begin{lemma}[Proposition 2.3 in \cite{seelmann2014notes}]\label{lemma, sin theta of infinite dimension operator}
Let $B$ be a self-adjoint operator on a separable Hilbert space $\widetilde{\mathcal{H}}$, and let ${V}\in\mathcal{L}(\widetilde{\mathcal{H}})$ be another self-adjoint operator where $\mathcal{L}\left(\widetilde{\mc H}\right)$ stands for the space of bounded linear operators from a Hilbert space $\widetilde{\mc H}$ to $\widetilde{\mc H}$.
Write the spectra of $B$ and $B+V$ as \[\mathrm{spec}( B)=\sigma\cup\Sigma\quad\text{and}\quad \mathrm{spec}( B+ V)=\omega\cup\Omega
\]
with $\sigma\cap\Sigma=\varnothing=\omega\cap\Omega$, and suppose that there is $\widehat d>0$ such that
\[\mathrm{dist}(\sigma,\Omega)\geqslant \widehat d\quad\text{and}\quad\mathrm{dist}(\Sigma,\omega)\geqslant \wh d\]
where $\mathrm dist(\sigma,\Sigma):=\min\{|a-b|:a\in\sigma,b\in\Omega\}$.
Then it holds that
\[\|P_{{B}}(\sigma)-P_{{B}+{V}}(\omega)\| \leqslant\frac\pi2\frac{\| V\| }{\wh d}\]
where $P_{ B}(\sigma)$ denotes the spectral projection for $ B$ associated with $\sigma$, i.e., 
\[P_{B}(\sigma):=\frac{1}{2\pi\mathrm{i}}\oint_{\gamma}\frac{\mathrm{d}z}{z-B},\]
where $\gamma$ is a contour on $\mathbb{C}$ that encloses $\sigma$ but no other elements of $\mathrm{spec}( B)$.
\end{lemma}
\begin{remark}
We note that, 
if further $ B$ is compact, 
the spectral projection coincide with projection operator onto the closure of the space spanned by the eigenfunctions associated with the eigenvalues in $\sigma$. 

Specifically, if $B$ is compact, by the spectral decomposition theorem one has
\[B=\sum_{i=1}^\infty\mu_ie_i\otimes e_i\quad\text{and}\quad(z- B)^{-1}=\sum_{i=1}^\infty(z-\mu_i)^{-1}e_i\otimes e_i,\]
where $\mr{spec}(B):=\{\mu_i\}_{i=1}^\infty$ satisfies $|\mu_i|\xrightarrow{i\to\infty} 0$.
Then $\forall v\in \mathcal{H}$, it holds that
\begin{align*}P_{B}(\sigma)v&=\frac{1}{2\pi\mathrm{i}}\oint_{\gamma}({z-B})^{-1}v~{\mathrm{d}z}=\frac{1}{2\pi\mathrm{i}}\oint_{\gamma}\sum_{i=1}^\infty(z-\mu_i)^{- 1}\langle e_i,v\rangle e_i~{\mathrm{d}z}\\
&=\sum_{i=1}^\infty\left[\left(\frac{1}{2\pi\mathrm{i}}\oint_{\gamma}(z-\mu_i)^{-1}~{\mathrm{d}z}\right)\langle e_i,v\rangle e_i\right]=\sum_{i\in\{i:\mu_i\in\sigma\}}\langle e_i,v\rangle e_i.
\end{align*}
In particular, if $\sigma=\mr{spec}(B)\backslash\{0\}$, then $P_{B}(\sigma)$ is the projection operator onto the $\overline{\mathrm{Im}}(B)$.
\end{remark}

Splitting eigenvalues into nonzero part and zero part yields the following useful corollary.
\begin{corollary}\label{cor: sin theta self adjoint}
Let $B$ and $B'$ be two positive semi-definite {and compact} operators with finite rank on a separable Hilbert space $\widetilde{\mathcal{H}}$. Let $\lambda_{\min}^+( B)$ and $\lambda_{\min}^+(B')$ be the infimum of the positive eigenvalues of ${B}$ and ${B}'$ respectively. Then we have
\[\left\|P_{ B}-P_{ B'}\right\| \leqslant\frac\pi2\frac{\| B- B'\| }{\min\{\lambda_{\min}^+( B),\lambda_{\min}^+( B')\}}.\]
\end{corollary}
\subsection{Sin Theta Theorem for General Operators}
When ${B}$ and ${V}$ in Lemma $\ref{lemma, sin theta of infinite dimension operator}$ are not self-adjoint, we use the symmetrization trick, which mainly depends on the following Lemma.
\begin{lemma}\label{lem:projection equality}
$P_A=P_{AA^*}$ for any bounded linear operator $A$ from a Hilbert space $\wt\H$ to $\wt\H$. Especially, $P_A=P_{AA^{\top}}$ for any matrix $A$.
\end{lemma}
\begin{proof}First we show that the null space of $A^*$ is the same as the null space of $AA^*$.
On the one hand, 
\[x\in\mathrm{null}(A^*)\Longrightarrow
A^*x=0\Longrightarrow AA^*x=0\Longrightarrow x\in\mathrm{null}(AA^*); 
\]
One the other hand,
\begin{align*}x\in\mathrm{null}(AA^*)&\Longrightarrow
AA^*x=0\Longrightarrow \langle x,AA^*x\rangle=\langle A^*x,A^*x\rangle=\|A^*x\|^2=0\\
&\Longrightarrow A^*x=0\Longrightarrow x\in\mathrm{null}(A^*).
\end{align*}
Hence, we have $\mathrm{null}(A^*)=\mathrm{null}(AA^*)$. Take the orthogonal complement of the both sides of this equality, we can get
\[\mathrm{null}(A^*)^{\perp}=\mathrm{null}(AA^*)^{\perp}\Longrightarrow {\mathrm{Im}(A)}={\mathrm{Im}(AA^*)}.\]
\end{proof}
Then we have the following Sin Theta theorem for general operator.
\begin{lemma}\label{lemma, sin theta of nonadjoint operator}
Let $ B,B'\in\mathcal{L}(\widetilde{\mathcal{H}})$ be two compact operators (not necessarily self-adjoint) with finite rank.
Then we have
\begin{align*}
\left\|P_{ B}-P_{ B'}\right\| &\leqslant\frac\pi2\frac{\| B B^*- B'B'^*\| }{\min\left\{\sigma_{\min}^+( B)^2,\sigma_{\min}^+(B')^2\right\}}\\
&\leqslant \frac\pi2\frac{\| B- B'\| ^2+2\| B- B'\| \| B'\| }{\min\left\{\sigma_{\min}^+( B)^2,\sigma_{\min}^+( B')^2\right\}}.
\end{align*}
\end{lemma}
\begin{proof}By Lemma $\ref{lem:projection equality}$, one can get $\left\|P_{ B}-P_{ B'}\right\| =\left\|P_{ B B^*}-P_{ B' B'^*}\right\| $.
Since $ BB^*, B'B'^*$ are both self-adjoint and compact, by Lemma $\ref{cor: sin theta self adjoint}$, one has
\begin{align*}
\left\|P_{ B B^*}-P_{ B' B'^*}\right\| \leqslant \frac{\pi}{2}\frac{\| B B^*- B' B'^*\| }{\min\left\{\lambda_{\min}^+\left( B B^*\right),\lambda_{\min}^+\left( B' B'^*\right)\right\}}.
\end{align*}
Then the proof is completed in view of the following inequality:
\begin{align}
\left\| B B^*- B' B'^*\right\| &= \|( B- B')( B- B')^*+( B-B')(B')^*+ B'( B- B')^*\| \nonumber\\
&\leqslant \| B- B'\| ^2+2\| B- B'\| \| B'\| . \label{eq:sy ineq}
\end{align}
\end{proof}

\section{Proof of Theorem \htheoremconv}
Thanks to the triangle inequality, one can bound the subspace estimation error by bounding the error term (i): $\mathbf{ Loss}_1:=\left\|P_{\mc S_{\Y|\X}^{(m)}}-P_{ \widehat {\mc S}_{\Y|\X}^{(m)}}\right\| $ and error term (ii): $\mathbf{ Loss}_2:= \left\|P_{\mathcal S_{\Y|\boldsymbol{X}}}-P_{\mathcal S_{\Y|\boldsymbol{X}}^{(m)}}\right\| $ respectively.
\subsection{Upper bound of error term (i)}
We first give the following lemmas, whose proofs are all deferred to the end of this section.
\begin{lemma}\label{lem:Gammam dagger Mm uniformly bounded}
Under Assumptions $\ref{as:joint distribution assumption}$ and $\ref{as:Linearity condition and Coverage condition}$, it holds that $\|\Gamma_m^\dagger M_m\| \leq \|\Gamma^{-1}M\| (\forall m).$
\end{lemma}
\begin{lemma}\label{lem: Gamma inverse M to Gammam dagger Mm}Under Assumptions $\ref{as:joint distribution assumption}$ and $\ref{as:Linearity condition and Coverage condition}$, we have \[\lim\limits_{m\to\infty}\lno\Gamma^{-1}M-\Gamma_m^\dagger M_m\rno =0.\]
\end{lemma}
\noindent We denote by $m_T(\varepsilon)$ the minimal integer $m_T$ satisfying $\lno\Gamma^{- 1}M-\Gamma_m^\dagger M_m\rno \leqslant \varepsilon$ for all $m\geqslant m_T$ and define an event 
$$\ttE:=\lb \left\|\widehat\Gamma_m^\dagger \widehat M_m^d-\Gamma_m^\dagger M_m\right\| \leqslant\mleft(\tfrac{D_0+1}{D_2}\mright)^{\frac52}\tfrac{24}{\wt C}n^{c_1(\alpha_1+1)+\gamma-\frac{1}{2}}+D_3n^{\frac{c_1(2\alpha_1+1)-1}{2}}\rb.$$
Then by taking $C$ to be $(D_0+1)n^{\frac{2\gamma}{5}}-\ln\l D_1m^2n \r$ in Proposition \ref{prop:bound of finite estimate}, one has: for $n\geqslant \l\frac{D_0+1}{D_2}\r^{\frac{5}{1-2\gamma}}$,
$$\P(\ttE)\geq 1-D_1m^2n\exp\left[-(D_0+1)n^{\frac{2\gamma}{5}}\right] -2\exp(- C'm).$$
\begin{lemma}\label{lem:lower bound sigma min total}
Introducing $
\bigtriangleup :=\max\lb \frac{\sigma_d(\Gamma^{-1} M)}{2},\frac{\sigma_d(\Gamma^{-1} M)^2}{4\|\Gamma^{-1}M\| } \rb$.
Suppose that Assumptions $\ref{as:joint distribution assumption}$ to $\ref{assumption: rate-type condition}$ hold, $c_1(2\alpha_1+1)-1<0$ and $2(c_1(\alpha_1+1)+\gamma)-1<0$. Then there exists a positive constant
\begin{align*}
n_2'=n_2'\l\sigma_d(\Gamma^{-1}M),\|\Gamma^{-1}M\| , \gamma,\sigma_0,\sigma_1,\bs K,m_M(1),c_1,m_T\l \tfrac{\bigtriangleup}{2}\r,\wt C,\alpha_1\r
\end{align*}
such that when $n\geqslant n_2'$, we have
\begin{align}
\sigma_{\min}^+(\Gamma_m^{\dagger} M_m)^2\geqslant \tfrac{\sigma_d(\Gamma^{-1}M)^2}{2} \label{eq: lower bound of sigma min}. 
\end{align}
Furthermore, Conditioning on $\ttE$, we have
\begin{align}\label{eq: lower bound of sigma min hat}
&\sigma_{\min}^+(\wh\Gamma_m^{\dagger}\wh M_m^d)^2\geqslant \tfrac{\sigma_d(\Gamma^{-1}M)^2}{2}.
\end{align}
\end{lemma}
The following proposition is an upper bound of error term (i):
\begin{proposition}\label{proposition, estimation error}
Positive constants $D_1$, $D_2$ and $C'$ as in Proposition $\ref{prop:bound of finite estimate}$,
suppose that Assumptions $\ref{as:joint distribution assumption}$ to $\ref{assumption: rate-type condition}$ hold, then $\forall \gamma\in(0,1/2)$, if $c_1$ satisfies $2c_1(\alpha_1+1)+2\gamma-1<0$ and $c_1(2\alpha_1+1)-1<0$, there exists a positive constant $C_1:=C_1\l \|\Gamma^{-1}M\| ,\sigma_d(\Gamma^{-1}M) ,\wt C,\gamma,\sigma_0,\sigma_1\r$ such that
\begin{align*}
\P\l
\lno P_{\mc{S}_{\Y|\X}^{(m)}}-P_{ \widehat{\mc{S}}_{\Y|\X}^{(m)}}\rno \leqslant C_1\frac{m^{\alpha_1+1}}{n^{1/2-\gamma}}\r\geqslant1-2\exp(- C'm)
- D_1m^2n\exp\l -(D_0+1)n^{\frac{2\gamma}{5}} \r&,
\end{align*}
when 
\begin{align*}
n\geqslant\max\Bigg\{ n_1,\l\tfrac{D_0+1}{D_2}\r^{\frac{5}{1-2\gamma}},\left[\tfrac{\|\Gamma^{-1}M\| \wt C}{48}\l\tfrac{D_2}{D_0+1}\r^{\frac52}\right]^{\frac{2}{2(c_1(\alpha_1+1)+\gamma)-1}}&,\\
\l \tfrac{\|\Gamma^{-1}M\| }{2D_3}\r^{\frac{2}{c_1(2\alpha_1+1)-1}},n_2',\left[ \tfrac{D_3\wt C}{24}\l \tfrac{D_2}{D_0+1} \r^{\frac52} \right]^{\frac2{2\gamma+c_1}}&\Bigg\}
\end{align*}
where $n_2'$ is defined in Lemma $\ref{lem:lower bound sigma min total}$.
\end{proposition}
\begin{proof}
By Lemma $\ref{lemma, way of estimate truncate central subspace}$, $\eqref{def: estimator central subspace}$ and Lemma $\ref{lemma, sin theta of nonadjoint operator}$, one has
\begin{align}
&\left\|P_{\mc S_{\Y|\vX}^{(m)}}-P_{\wh{\mc{S}}_{\Y|\vX}^{(m)}}\right\| =\left\|P_{\Gamma_m^{\dagger}M_m}-P_{\wh\Gamma_m^{\dagger}\wh M_m^d}\right\| \nonumber\\
&\qquad\leqslant\frac{\pi}{2}\frac{\lno\widehat\Gamma_m^\dagger \widehat M_m^d-\Gamma_m^\dagger M_m\rno ^2+\lno\widehat\Gamma_m^\dagger \widehat M_m^d-\Gamma_m^\dagger M_m\rno \lno\Gamma_m^\dagger M_m\rno }{\min\lb\sigma_{\min}^+\l\wh\Gamma_m^\dagger \wh M_m^d\r^2,\sigma_{\min}^+\l\Gamma_m^\dagger M_m\r^2\rb}\label{eq: PS minus P hat S norm}.
\end{align}
Because of $c_1(2\alpha_1+1)-1<0$ and $2(c_1(\alpha_1+1)+\gamma)-1<0$, it is easy to check that when
\[n\geqslant\max\lb\left[\tfrac{\|\Gamma^{-1}M\| \wt C}{48}\l\tfrac{D_2}{D_0+1}\r^{\frac52}\right]^{\frac{2}{2(c_1(\alpha_1+1)+\gamma)-1}},\l \tfrac{\|\Gamma^{-1}M\| }{2D_3}\r^{\frac{2}{c_1(2\alpha_1+1)-1}}\rb,\]
both $\l\tfrac{D_0+1}{D_2}\r^{\frac52}\tfrac{24}{\wt C}n^{c_1(\alpha_1+1)+\gamma-\frac{1}{2}}$ and $D_3n^{\frac{c_1(2\alpha_1+1)-1}{2}}$ are less than or equal to $\frac{\|\Gamma^{-1}M\| }{2}$. Thus, on the event $\ttE$,
\begin{align}\label{eq: high prob upper bound is Gamma minus 1 M}
\lno\widehat\Gamma_m^\dagger \widehat M_m^d-\Gamma_m^\dagger M_m\rno \leqslant \lno\Gamma^{-1}M\rno .
\end{align}
By Lemma $\ref{lem:Gammam dagger Mm uniformly bounded}$, inserting \eqref{eq: high prob upper bound is Gamma minus 1 M} into \eqref{eq: PS minus P hat S norm} leads to
$$
\lno P_{\mc{S}_{\Y|\X}^{(m)}}-P_{ \widehat{\mc{S}}_{\Y|\X}^{(m)}}\rno
\leqslant \frac{\pi\lno\widehat\Gamma_m^\dagger \widehat M_m^d-\Gamma_m^\dagger M_m\rno \lno\Gamma^{-1}M\rno }{\min\lb\sigma_{\min}^+\l\wh\Gamma_m^\dagger \wh M_m^d\r^2,\sigma_{\min}^+\l\Gamma_m^\dagger M_m\r^2\rb},
$$
on the event $\ttE$.
Furthermore, when $n\geqslant \left[ \frac{D_3\wt C}{24}\l \frac{D_2}{D_0+1} \r^{\frac52} \right]^{\frac2{2\gamma+c_1}}$ and $n\geq n_2'$, one can get
$\l \tfrac{D_0+1}{D_2}\r^{\frac52}\tfrac{24m^{\alpha_1+1}}{\wt C n^{1/2-\gamma}}$ is greater than or equal to $D_3\tfrac{m^{(2\alpha_1+1)/2}}{n^{1/2}}$
and then on the event $\ttE$,
\begin{align*}
\lno P_{\mc{S}_{\Y|\X}^{(m)}}-P_{ \widehat{\mc{S}}_{\Y|\X}^{(m)}}\rno \leqslant \tfrac{96\pi\|\Gamma^{-1}M\| }{\sigma_d(\Gamma^{-1}M)^2}\l \tfrac{D_0+1}{D_2}\r^{\frac52}\tfrac{m^{\alpha_1+1}}{\wt C n^{1/2-\gamma}}.
\end{align*}
 by
Lemma $\ref{lem:lower bound sigma min total}$.
Then choosing $C_1=\tfrac{96\pi\|\Gamma^{-1}M\| }{\wt C\sigma_d(\Gamma^{-1}M)^2}\l \tfrac{D_0+1}{D_2}\r^{\frac52}$ can complete the proof.
\end{proof}

\begin{proof}[\textbf{Proof of Lemma \ref{lem:Gammam dagger Mm uniformly bounded}}]
First, it is easy to check that:
\begin{align}
\Gamma^\dag_m=\Pi_m\Gamma^{-1}\Pi_m=\Pi_m\Gamma^{-1}=\Gamma^{-1}\Pi_m=\sum\limits_{i=1}^m\lambda_i^{-1}\phi_i\otimes\phi_i.\label{eq: Gamma m dag def}
\end{align}
According to \eqref{eq: Gamma m dag def} and $M_m=\Pi_mM\Pi_m$, it is easy to check that $\Gamma_m^\dagger M_m=\Pi_m \Gamma^{- 1}M\Pi_m$. Then by the compatibility of operator norm, one can get
\begin{align*}
\lno\Gamma_m^\dagger M_m\rno =\lno\Pi_m \Gamma^{-1}M\Pi_m\rno \leqslant \lno\Pi_m\rno \lno\Gamma^{-1}M\rno \lno\Pi_m\rno =\lno\Gamma^{-1}M\rno .
\end{align*}
Note that $\Gamma^{-1}M$ is bounded since $\Gamma^{-1}M$ is of finite rank by Corollary $\ref{corollary, MDDO and central subspace}$. Thus the proof is completed. 
\end{proof}

\begin{proof}[\textbf{Proof of Lemma \ref{lem: Gamma inverse M to Gammam dagger Mm}}]
It is easy to check that
$\Gamma_m^\dagger M_m=\Pi_m\Gamma^{-1}M\Pi_m$ and $\Gamma^{-1}M$ is of finite rank by Corollary $\ref{corollary, MDDO and central subspace}$.
Thus the proof is completed by Lemma $\ref{lem:PimTPimtoT}$.
\end{proof}
\begin{proof}[\textbf{Proof of Lemma \ref{lem:lower bound sigma min total}}]
We first prove \eqref{eq: lower bound of sigma min}.
By Corollary $\ref{corollary, MDDO and central subspace}$ and Lemma $\ref{lem:projection equality}$, one has $\rank(\Gamma^{- 1}M)=\rank\l\Gamma^{- 1}M(\Gamma^{- 1}M)^*\r=d$. Thus
\begin{align*}
\sigma_{\min}^+(\Gamma^{-1}M)^2=\lambda_{\min}^+\l\Gamma^{-1}M(\Gamma^{-1}M)^*\r=\lambda_d\l \Gamma^{-1}M(\Gamma^{-1}M)^*\r. 
\end{align*}
 It is easy to see $\rank(\Gamma_m^\dagger M_m)=\rank\l \Gamma_m^\dagger M_m(\Gamma_m^\dagger M_m)^*\r\leqslant d$ by $\Gamma_m^\dagger M_m=\Pi_m \Gamma^{-1} M \Pi_m$ and Lemma $\ref{lem:projection equality}$, thus one can assume that 
 \begin{align*}
\sigma_{\min}^+(\Gamma^\dagger_m M_m)^2=\lambda_{\min}^+\l\Gamma_m^\dagger M_m(\Gamma_m^\dagger M_m)^*\r=\lambda_j\l \Gamma_m^\dagger M_m(\Gamma_m^\dagger M_m)^*\r
\end{align*}
for some $j\leqslant d$.
By Corollary $\ref{coro:wely ineq operator}$, $\eqref{eq:sy ineq}$ and
Lemma $\ref{lem: Gamma inverse M to Gammam dagger Mm}$
, one has
\begin{align*}
\left|\sigma_{\min}^+(\Gamma^\dagger_m M_m)^2-\sigma_j(\Gamma^{-1} M)^2\right|&=\left|\lambda_{j}\l\Gamma^\dagger_m M_m(\Gamma^\dagger_m M_m)^{*}\r-\lambda_j\l \Gamma^{-1} M(\Gamma^{-1} M)^*\r\right|\\
&\leqslant
\|\Gamma^{-1} M(\Gamma^{-1} M)^*- \Gamma_m^\dagger M_m(\Gamma_m^\dagger M_m)^*\| \\
&\leqslant \|\Gamma^{-1} M- \Gamma_m^\dagger M_m\| ^2+
\|\Gamma^{-1} M- \Gamma_m^\dagger M_m\| \cdot\|\Gamma^{-1} M\| \xrightarrow{m\to\infty} 0. 
\end{align*}
Thus, for \[
n\geqslant m_T(\bigtriangleup)^{\frac1{c_1}}=m_T\l\max\lb\tfrac{\sigma_d(\Gamma^{-1} M)}{2},\tfrac{\sigma_d(\Gamma^{-1} M)^2}{4\|\Gamma^{-1}M\| }\rb\r^{\frac1{c_1}}, 
\]
one has $\|\Gamma^{-1} M- \Gamma_m^\dagger M_m\| ^2$ and $\|\Gamma^{-1} M- \Gamma_m^\dagger M_m\| \cdot\|\Gamma^{-1} M\|$ are both less than or equal to $\frac{1}{4}\sigma_d(\Gamma^{-1} M)^2$. Hence one can get
$\left|\sigma_{\min}^+(\Gamma^\dagger_m M_m)^2-\sigma_j(\Gamma^{-1} M)^2\right|\leqslant\frac{1}{2}\sigma_d(\Gamma^{-1} M)^2$
and
\begin{equation}
\sigma_{\min}^+(\Gamma^\dagger M_m)^2\geqslant \sigma_j(\Gamma^{-1} M)^2-\frac{1}{2}\sigma_d(\Gamma^{-1} M)^2\geqslant\frac{1}{2}\sigma_d(\Gamma^{-1} M)^2
\end{equation}
for sufficiently large $n$. This completes the proof of \eqref{eq: lower bound of sigma min}.

Next, we prove $\eqref{eq: lower bound of sigma min hat}$. Combining Proposition $\ref{prop:bound of finite estimate}$ with Lemma $\ref{lem: Gamma inverse M to Gammam dagger Mm}$ leads to that on the event $\ttE$, 
\[\lno\wh \Gamma_m^\dag\wh M^d_m- \Gamma^{-1}M\rno \leqslant\ve+\l\tfrac{D_0+1}{D_2}\r^{\frac52}\tfrac{24}{\wt C}n^{c_1(\alpha_1+1)+\gamma-\frac{1}{2}}+D_3n^{\frac{c_1(2\alpha_1+1)-1}{2}}
\]
holds for $n\geqslant \max\{n_1,m_T( \ve)^{1/c_1}\}$.
Assuming that $c_1(2\alpha_1+1)-1<0$ and $2(c_1(\alpha_1+1)+\gamma)-1<0$, it is easy to check that when 
\[n\geqslant\max\lb\left[\tfrac{\bigtriangleup \wt C}{96}\l\tfrac{D_2}{D_0+1}\r^{\frac52}\right]^{\frac{2}{2(c_1(\alpha_1+1)+\gamma)-1}},\l \tfrac{\bigtriangleup}{4D_3}\r^{\frac{2}{c_1(2\alpha_1+1)-1}}\rb,\] both $\l\tfrac{D_0+1}{D_2}\r^{\frac52}\tfrac{24}{\wt C}n^{c_1(\alpha_1+1)+\gamma-\frac{1}{2}}$ and $D_3n^{\frac{c_1(2\alpha_1+1)-1}{2}}$ are less than or equal to $\frac{\bigtriangleup}{4}$. Letting $\varepsilon=\frac12\bigtriangleup$, one can get on the event $\ttE$,
when
\begin{align*}
n&\geqslant n_2'=n_2'\l\sigma_d(\Gamma^{-1}M),\|\Gamma^{-1}M\| , \gamma,\sigma_0,\sigma_1,\bs K,m_M(1),c_1,m_T\l \tfrac{\bigtriangleup}{2}\r,\wt C,\alpha_1\r\\
&:=\max\bigg\{ n_1,m_T\l \tfrac{\bigtriangleup}{2}\r^{1/c_1}, \left[\tfrac{\bigtriangleup \wt C}{96}\l\tfrac{D_2}{D_0+1}\r^{\frac52}\right]^{\frac{2}{2(c_1(\alpha_1+1)+\gamma)-1}},\l \tfrac{\bigtriangleup}{4D_3}\r^{\frac{2}{c_1(2\alpha_1+1)-1}}\bigg\},
\end{align*}
one has $\lno\wh \Gamma_m^\dag\wh M^d_m- \Gamma^{-1}M\rno \leqslant\bigtriangleup$ and further
$\sigma_{\min}^+(\wh\Gamma^\dagger \wh M^d_m)^2\geqslant \tfrac{\sigma_d(\Gamma^{-1} M)^2}{2}$ by the same argument as the proof of \eqref{eq: lower bound of sigma min}.
 This completes the proof of \eqref{eq: lower bound of sigma min hat}.
Considering that $m_T(\bigtriangleup)\leqslant m_{T}\l\frac\bigtriangleup2\r$, one can also get $\eqref{eq: lower bound of sigma min}$ when $n\geqslant n_2'$. Thus the proof is completed.
\end{proof}
\subsection{Upper bound of error term (ii)}\label{ap, subs, truncation error}
\begin{proposition}\label{proposition, truncation error}
Under Assumption $\ref{assumption: rate-type condition}$, there exists a positive constant $C_2:=C_2\l d,\wt C,\lambda_d(\mc{B}),\alpha_2\r$ where $\mc{B}:=\sum\limits_{i=1}^d {\bs{\beta}}_i\otimes{\bs{\beta}}_i$ for ${\bs{\beta}}_i$ defined in \eqref{def: central subspace}, such that when $n\geqslant \l \frac{\lambda_d({\mc{B}})}{4d\wt C^2}\sqrt{\frac{2\alpha_2-1}{\zeta(2\alpha_2)}}\r^{\frac{2}{c_1(1-2\alpha_2)}}$, we have
\begin{equation}\label{equation, truncation error}
 \left\|P_{\mathcal S_{\Y|\boldsymbol{X}}}-P_{\mathcal S_{\Y|\boldsymbol{X}}^{(m)}}\right\| \leqslant C_2m^{-\frac{2\alpha_2-1}{2}},
\end{equation}
where $\zeta(\cdot)$ is Riemann $\zeta$ function.
\end{proposition}
\begin{proof}
Let ${\mc{B}^{(m)}}:=\sum\limits_{i=1}^d {\bs{\beta}}_i^{(m)}\otimes{\bs{\beta}}_i^{(m)}$.
Combing with Equation $\eqref{def: central subspace}$, it is easy to check that $\left\|P_{\mathcal S_{\Y|\boldsymbol{X}}}-P_{\mathcal S_{\Y|\boldsymbol{X}}^{(m)}}\right\| =\|P_{\mc{B}}-P_{\mc{B}^{(m)}}\| $. By Corollary $\ref{cor: sin theta self adjoint}$, we have
\begin{align}\label{eq:sin theta for B Bm}
\|P_{\mc{B}}-P_{\mc{B}^{(m)}}\| \leqslant \frac{\pi}{2}\frac{\|{\mc{B}}-{\mc{B}^{(m)}}\| }{\min\{\lambda_{\min}^+({\mc{B}}),\lambda_{\min}^+({\mc{B}^{(m)}})\}}.
\end{align}

Note that ${\mc{B}}-{\mc{B}^{(m)}}$ is self-adjoint, then
\begin{align*}
&\lno{\mc{B}}-{\mc{B}^{(m)}}\rno =\sup_{{\bs{\beta}}\in\mathbb{S}_{ \mathcal H}}|\langle ({\mc{B}}-{\mc{B}^{(m)}})({\bs{\beta}}),{\bs{\beta}}\rangle|=\sup_{{\bs{\beta}}\in\mathbb{S}_{\mathcal H}}|\langle {\mc{B}}{\bs{\beta}},{\bs{\beta}}\rangle-\langle {\mc{B}^{(m)}}{\bs{\beta}},{\bs{\beta}}\rangle|\\
&~~=\sup_{{\bs{\beta}}\in\mathbb{S}_{\mathcal H}}\left|\sum_{i=1}^d\left[\langle{\bs{\beta}}_i,{\bs{\beta}}\rangle^2-\langle{\bs{\beta}}_i^{(m)},{\bs{\beta}}\rangle^2\right]\right|=\sup_{{\bs{\beta}}\in\mathbb{S}_{\mathcal H}}\left| \sum_{i=1}^d\langle{\bs{\beta}}_i-{\bs{\beta}}_i^{(m)},{\bs{\beta}}\rangle\langle{\bs{\beta}}_i+{\bs{\beta}}_i^{(m)},{\bs{\beta}}\rangle\right|\\
&~~\leqslant\sup_{{\bs{\beta}}\in\mathbb{S}_{\mathcal H}}\sum_{i=1}^d\left| \langle{\bs{\beta}}_i-{\bs{\beta}}_i^{(m)},{\bs{\beta}}\rangle\langle{\bs{\beta}}_i+{\bs{\beta}}_i^{(m)},{\bs{\beta}}\rangle\right|
\leqslant\sum_{i=1}^d\left\|{\bs{\beta}}_i-{\bs{\beta}}_i^{(m)}\right\|\left\|{\bs{\beta}}_i+{\bs{\beta}}_i^{(m)}\right\|,
\end{align*}
where the first inequality comes from the triangle inequality, and the 
second inequality comes from the Cauchy-Schwarz inequality and $\|{\bs{\beta}}\|=1$. 
 Then one has ${\bs{\beta}}_i=\sum\limits_{j=1}^\infty b_{ij}\phi_j$ and 
\[{\bs{\beta}}^{(m)}_i=\Pi_m{\bs{\beta}}_i=\sum_{j'=1}^m\phi_{j'}\otimes\phi_{j'}\sum_{j=1}^\infty b_{ij}\phi_j=\sum_{j'=1}^m\sum_{j=1}^\infty\langle\phi_{j'},\phi_j\rangle b_{ij}\phi_{j'}=\sum_{j=1}^mb_{ij}\phi_j.\]
According to Assumption $\ref{assumption: rate-type condition}$, one can get
\begin{align*}
\left\|{\bs{\beta}}_i-{\bs{\beta}}_i^{(m)}\right\|&=\left\|\sum_{j=m+1}^\infty b_{ij}\phi_j\right\|=\sqrt{\sum_{j=m+1}^\infty b_{ij}^2}\leqslant \wt C\sqrt{\sum_{j=m+1}^\infty j^{-2\alpha_2}};\\
\left\|{\bs{\beta}}_i+{\bs{\beta}}_i^{(m)}\right\|&\leqslant\|{\bs{\beta}}_i\|+\lno{\bs{\beta}}_i^{(m)}\rno\leqslant2\|{\bs{\beta}}_i\|=2\sqrt{\sum_{j=1}^\infty b_{ij}^2}\leqslant 2\wt C\sqrt{\sum_{j=1}^\infty j^{- 2\alpha_2}}.
\end{align*}
Because $\alpha_2>1/2$, one has
\[\sum\limits_{j=m+1}^\infty \frac{1}{j^{2\alpha_2}}\leqslant \frac{1}{2\alpha_2-1}\frac{1}{m^{2\alpha_2-1}};\qquad \sum_{j=1}^\infty \frac 1{j^{2\alpha_2}}=\zeta(2\alpha_2)\text{ is convergent},\]
where $\zeta(\cdot)$ is Riemann $\zeta$ function. Thus, one can get
\begin{equation}\label{eq: upper bound of operator norm of A minus B}
\lno{\mc{B}}-{\mc{B}^{(m)}}\rno \leqslant 2d\wt C^2\sqrt{\frac{\zeta(2\alpha_2)}{2\alpha_2-1}}m^{-\frac{2\alpha_2-1}{2}}.
\end{equation}

Furthermore, 
{since $\mr{rank}(\mc{B})=d$, one can get that $\lambda_{\min}^+(\mc{B})=\lambda_{d}(\mc{B})$. It is easy to see $\rank(\mc{B}^{(m)})\leqslant d$ by $\mc{B}^{(m)}=\Pi_m \mc{B} \Pi_m$, thus one can assume that $\lambda_{\min}^+(\mc{B}^{(m)})=\lambda_j( \mc{B}^{(m)})$ for some $j\leqslant d$.
By Corollary $\ref{coro:wely ineq operator}$
and \eqref{eq: upper bound of operator norm of A minus B}, one has:
$$
|\lambda_j( \mc{B}^{(m)})-\lambda_j\l \mc{B}\r|\leqslant\lno \mc{B}-\mc{B}^{(m)}\rno \leqslant 2d\wt C^2\sqrt{\frac{\zeta(2\alpha_2)}{2\alpha_2-1}}m^{-\frac{2\alpha_2-1}{2}}.
$$
Thus for sufficiently large {$n\geqslant \l \frac{\lambda_d({\mc{B}})}{4d\wt C^2}\sqrt{\frac{2\alpha_2-1}{\zeta(2\alpha_2)}} \r^{\frac{2}{c_1(1-2\alpha_2)}}$}, one has
\begin{align}
&\lambda_j\l \mc{B}^{(m)}\r\geqslant \lambda_j\l \mc{B}\r-\frac{\lambda_d\l \mc{B}\r}{2}
\geqslant\frac{\lambda_d\l \mc{B}\r}{2}\nonumber\\
&\qquad\Longrightarrow \min\{\lambda_{\min}^+({\mc{B}}),\lambda_{\min}^+({\mc{B}^{(m)}})\}\geqslant \frac{\lambda_d({\mc{B}})}{2}. \label{eq:lower bound lambda min plus B Bm}
\end{align}}
Inserting \eqref{eq: upper bound of operator norm of A minus B} and \eqref{eq:lower bound lambda min plus B Bm} into \eqref{eq:sin theta for B Bm} leads to
\begin{align*}
\left\|P_{\mathcal S_{\Y|\boldsymbol{X}}}-P_{\mathcal S_{\Y|\boldsymbol{X}}^{(m)}}\right\| \leqslant \frac{2\pi d\wt C^2}{\lambda_{d}(\mc{B})}\sqrt{\frac{\zeta(2\alpha_2)}{2\alpha_2-1}}m^{-\frac{2\alpha_2-1}{2}}.
\end{align*}
Then choosing $C_2:=\frac{2\pi d\wt C^2}{\lambda_d({\mc{B}})}\sqrt{\frac{\zeta(2\alpha_2)}{2\alpha_2-1}}$ can complete the proof.
\end{proof}

\subsection{Proof of Theorem \htheoremconv}
\begin{proof}
Note that
\begin{equation}
\begin{aligned}
\left\|P_{\mc{S}_{\Y|\X}}-P_{\widehat{\mc{S}}_{\Y|\X}^{(m)}}\right\| 
&\leqslant \left\|P_{\mc{S}_{\Y|\X}}-P_{\mc{S}_{\Y|\X}^{(m)}}\right\| +\left\|P_{\mc{S}_{\Y|\X}^{(m)}}-P_{ \widehat{\mc{S}}_{\Y|\X}^{(m)}}\right\| .\\
\end{aligned}
\end{equation}
Next we select $m$ to be $n^{\frac{1-2\gamma}{2\alpha_1+2\alpha_2+1}}$, i.e., $c_1:=\frac{1-2\gamma}{2\alpha_1+2\alpha_2+1}$. And it is easy to check that $c_1$ satisfies $2c_1(\alpha_1+1)+2\gamma-1=-\frac{(1-2\gamma)(2\alpha_2-1)}{2\alpha_1+2\alpha_2+1}<0$ and $c_1(2\alpha_1+1)-1=-\frac{2[\gamma(2\alpha_1+1)+\alpha_2]}{2\alpha_1+2\alpha_2+1}<0$.
Then combining Proposition $\ref{proposition, estimation error}$ with Proposition $\ref{proposition, truncation error}$ leads to
\begin{align*}
\P\mpt{\left\|P_{\mc S_{\Y|\X}}-P_{\widehat{\mc{S}}_{\Y|\X}^{(m)}}\right\| \leqslant (C_1+C_2)n^{-\frac{(2\alpha_2-1)(1-2\gamma)}{2(2\alpha_1+2\alpha_2+1)}}}\geqslant 1-2\exp\l- C'n^{\frac{1-2\gamma}{2\alpha_1+2\alpha_2+1}}\r&\\
-\exp\left[\ln\l D_1n^{\frac{2\alpha_1+2\alpha_2+3-4\gamma}{2\alpha_1+2\alpha_2+1}} \r-(D_0+1)n^{\frac{2\gamma}{5}}\right]&
\end{align*}
when $n\geqslant n_3'$, where
\begin{align*}
n_3'=\max\Bigg\{n_1,n_2',\left[\tfrac{\|\Gamma^{-1}M\| \wt C}{48}\l\tfrac{D_2}{D_0+1}\r^{\frac52}\right]^{\frac{2}{2(c_1(\alpha_1+1)+\gamma)-1}},\l \tfrac{\|\Gamma^{-1}M\| }{2D_3}\r^{\frac{2}{c_1(2\alpha_1+1)-1}},\\
\l\tfrac{D_0+1}{D_2}\r^{\frac{5}{1-2\gamma}},\left[ \tfrac{D_3\wt C}{24}\l \tfrac{D_2}{D_0+1} \r^{\frac52} \right]^{\frac2{2\gamma+c_1}},\l\tfrac{\lambda_d(\mc{B})}{4d\wt C^2}\sqrt{\tfrac{{2\alpha_2-1}}{\zeta(2\alpha_2)}} \r^{\frac{2}{c_1(1-2\alpha_2)}}\Bigg\}.
\end{align*}
It is easy to check that if $\frac{2\gamma}{5}<\frac{1-2\gamma}{2\alpha_1+2\alpha_2+1}$, i.e., $\gamma<\frac{5}{4(\alpha_1+\alpha_2+3)}$ holds, 
there exists a constant $n_3''=n_3''\l \gamma,\alpha_1,\alpha_2,D_0,D_1,C'\r$ such that when $n\geqslant n_3'$ further, we have 
\begin{align*}
\P\l\left\|P_{\mc S_{\Y|\X}}-P_{\widehat{\mc{S}}_{\Y|\X}^{(m)}}\right\| \leqslant (C_1+C_2)n^{-\frac{(2\alpha_2-1)(1-2\gamma)}{2(2\alpha_1+2\alpha_2+1)}} \r
\geqslant1-2\exp\l-\tfrac{D_0+1}{2}n^{\frac{2\gamma}{5}} \r.
\end{align*}
Thus, one can choose $n_3=\max\{n_3',n_3''\}$ to get the following conclusion.
\begin{proposition}
Under Assumptions $\ref{as:joint distribution assumption}$ to $\ref{assumption: rate-type condition}$, for any $\gamma\in\l0,\tfrac{5}{4(\alpha_1+\alpha_2+3)}\r$, choosing 
$m=n^{\frac{1-2\gamma}{2\alpha_1+2\alpha_2+1}}$ (i.e., $c_1=\frac{1-2\gamma}{2\alpha_1+2\alpha_2+1}$) yields a positive constant
\begin{align*}
D_4:=D_4\mpt{ \|\Gamma^{-1}M\| ,\sigma_d(\Gamma^{-1}M) ,\gamma,\sigma_0,\sigma_1,d,\wt C,\lambda_d\big({\textstyle\sum\limits_{i=1}^d {\bs{\beta}}_i\otimes{\bs{\beta}}_i}\big),\alpha_2} 
\end{align*}
such that when $n$ is sufficiently large, we have:
\begin{align*}
\P\l\left\|P_{\mc{S}_{\Y|\X}}-P_{\widehat{\mc{S}}_{\Y|\X}^{(m)}}\right\| \leqslant D_4n^{-\frac{(2\alpha_2-1)(1-2\gamma)}{2(2\alpha_1+2\alpha_2+1)}} \r
\geqslant1-2\exp\l -\tfrac{D_0+1}{2}n^{\frac{2\gamma}{5}} \r,
\end{align*}
where $D_0$ and $D_1$ are defined in Proposition $\ref{prop:bound hatMmd Mm}$.
\end{proposition}
\noindent
Define 
$$\mathtt F:=\left\{\left\|P_{\mc{S}_{\Y|\X}}-P_{\widehat{\mc{S}}_{\Y|\X}^{(m)}}\right\| \leqslant D_4n^{-\frac{(2\alpha_2-1)(1-2\gamma)}{2(2\alpha_1+2\alpha_2+1)}}\right\}.$$
Then 
\begin{align*}
 &\mb E\left[\left\|P_{\mc{S}_{\Y|\X}}-P_{\widehat{ \mc{S}}_{\Y|\X}^{(m)}}\right\|^2\right] =
 \mb E\left[\left\|P_{\mc{S}_{\Y|\X}}-P_{\widehat{ \mc{S}}_{\Y|\X}^{(m)}}\right\|^21_{\mathtt{F}}\right] +
 \mb E\left[\left\|P_{\mc{S}_{\Y|\X}}-P_{\widehat{ \mc{S}}_{\Y|\X}^{(m)}}\right\|^21_{\mathtt{F}^c}\right]\\ 
&\qquad \leqslant 
 D_4^2n^{-\frac{(2\alpha_2-1)(1-2\gamma)}{2\alpha_1+2\alpha_2+1}}+4\mb P\left( \mathtt F^c\right)
 \lesssim n^{-\frac{(2\alpha_2-1)(1-2\gamma)}{2\alpha_1+2\alpha_2+1}}+\exp\l -\tfrac{D_0+1}{2}n^{\frac{2\gamma}{5}} \r
\lesssim n^{-\frac{(2\alpha_2-1)(1-2\gamma)}{2\alpha_1+2\alpha_2+1}}.
\end{align*}
This completes the proof of Theorem \ref{theorem, total convergence rate}.
\end{proof}





\end{document}